\documentclass[romanappendices,journal,compsoc]{IEEEtran}

\usepackage[numbers]{natbib}
\usepackage{times}
\usepackage{subfigure}
\usepackage{graphicx}
\usepackage{amsmath,amssymb}
\usepackage{bbold}
\usepackage[breaklinks=true,colorlinks,bookmarks=false]{hyperref}
\usepackage{ragged2e}
\usepackage{multirow}
\usepackage[table,xcdraw]{xcolor}
\usepackage[accsupp]{axessibility}  % Improves PDF readability for those with disabilities.
\usepackage{orcidlink}

\graphicspath{{Figures/}}

\begin{document}

\title{Context-Aware Mixture of Domain Experts for Bodily Expression of Emotion in the Wild}

\author{Mohammad Mahdi~Dehshibi~\orcidlink{0000-0001-8112-5419}~\IEEEmembership{Senior~Member, IEEE}\thanks{\upshape Department of Computer Science, University of the West of England, Bristol, U.K. Email: mohammad.dehshibi@uwe.ac.uk} and David~Masip~\orcidlink{0000-0001-7898-1847}~\IEEEmembership{Senior~Member, IEEE}\thanks{\upshape \\ Department of Computer Science, Universitat Oberta de Catalunya, Barcelona, Catalonia, Spain. Email: dmasipr@uoc.edu}}

\IEEEtitleabstractindextext{
\justify
\begin{abstract}
The same body posture can convey entirely different emotions depending on its surrounding context, yet most methods for recognising bodily emotions treat scene and object cues as auxiliary feature augmentations rather than as structured priors over the plausibility of emotions. We introduce the Context-Aware Mixture of Domain Experts (CA-MoDE) for bodily emotion recognition. CA-MoDE incorporates dedicated scene and object experts to generate soft distributions over emotion categories conditioned on their respective domains. These domain-conditioned soft predictions serve as structured contextual priors that modulate the body expert's predictions at the distributional level rather than at the feature level. To fuse these multi-domain signals, we propose a task-tailored max-endorsement gating strategy that selects the strongest contextual signal across experts for each emotion dimension. Our gating strategy mitigates the signal dilution that typically occurs when conflicting or uninformative context distributions are averaged. CA-MoDE achieves an Emotion Recognition Score of 0.3269 on the Body Language Database. By outperforming existing temporal models using only single still images, our framework demonstrates that explicitly modelling structured spatial context can serve as a complementary discriminative proxy for the behavioural dynamics typically captured by video. 

\end{abstract}

\begin{IEEEkeywords}
Bodily Emotion Recognition, Context-aware Learning, Max-Endorsement Gating, Mixture of Domain Experts, Co-occurrence Prior.
\end{IEEEkeywords}}

%=============SECTION I==================
\maketitle
\begin{figure}[!htbp]
    \centering
    \includegraphics[width=\linewidth]{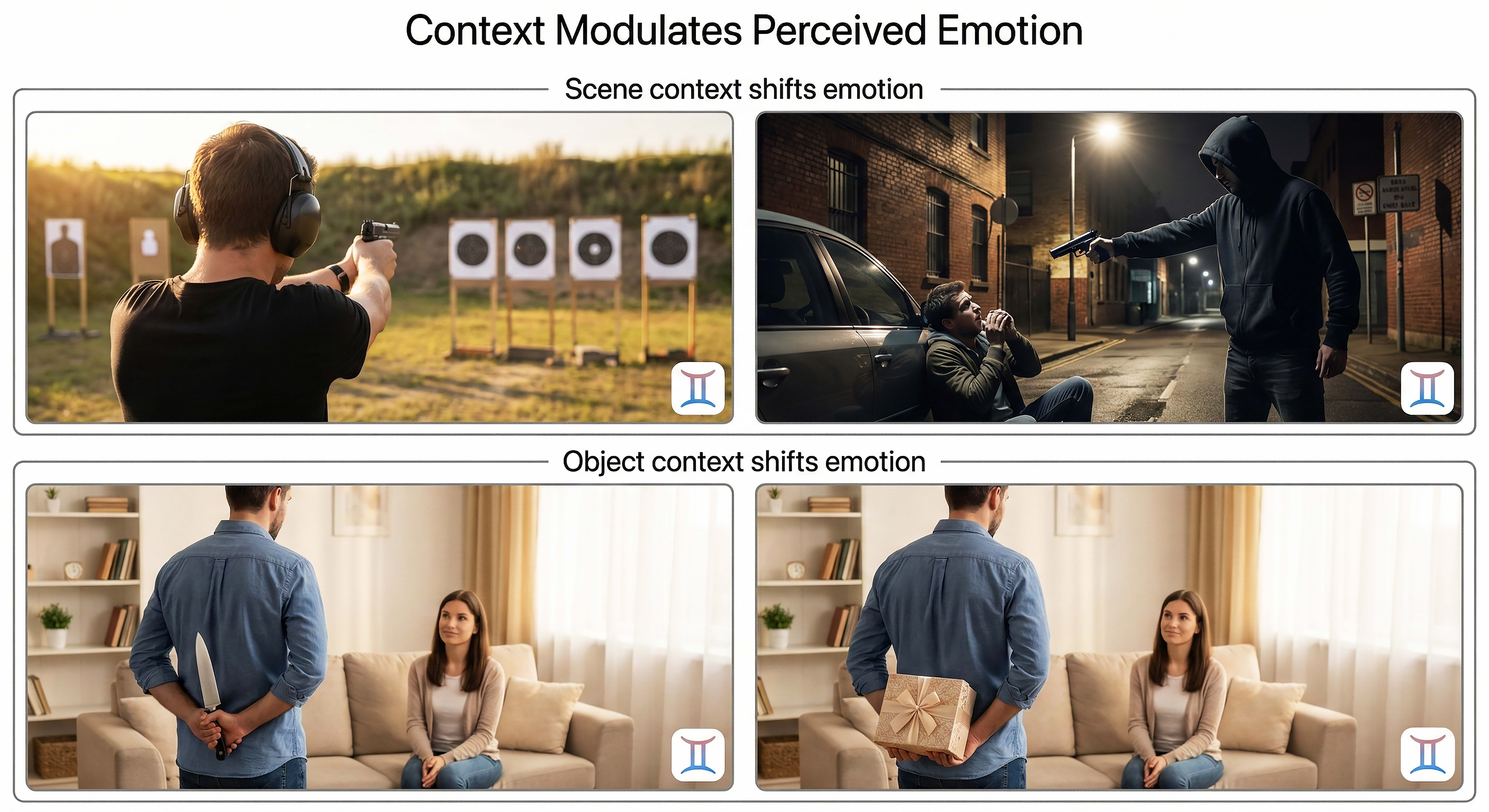}
    \caption{Context fundamentally alters the perceived emotion of an action. The same object (a handgun) evokes pleasure and excitement in a shooting range, but fear and violence on a street. The same scene (a living room) evokes threat or affection depending on whether a partner holds a knife or a gift box. \textit{These synthetic images are used for illustrative purposes only and do not depict real individuals or events.}}
    \label{fig:teaser}
\end{figure}

\IEEEraisesectionheading{\section{Introduction}
\label{sec:introduction}}
\IEEEPARstart{E}{motion} is not a property of the body in isolation. The same posture can signal excitement or fear, aversion or affection, depending on the surrounding scene and objects. Psychological studies confirm that body language is systematically ambiguous without its surrounding context~\cite{aviezer2012body}, and that the environment in which a gesture occurs can alter its emotional interpretation~\cite{dadamo2024soniweight,dehshibi2021deep,dehshibi2023pain,hens2025last,kleinsmith2011automatic,ortigoso2025lightweight}. Recognising bodily expressions of emotion in the wild, therefore, requires reasoning about context, not just about the body. Figure~\ref{fig:teaser} illustrates this dependence of bodily emotion on scene and object context.

%------PLACE

Automatic Identification of Bodily Expression of Emotions (AIBEE) has advanced substantially since the introduction of large-scale benchmarks such as EMOTIC~\cite{kosti2020context} and BoLD~\cite{luo2020arbee}. However, this progress has largely improved representation learning rather than changing how contextual evidence is modelled. In most architectures, contextual cues (e.g., scene categories and detected objects) are appended as auxiliary feature channels to a body-based backbone~\cite{filntisis2020emotion,9666957}. The network is then expected to discover that, for instance, a detected weapon shifts the probability distribution from pleasure to fear. This design has a predictable failure mode. When scene context provides a strong signal and object context is silent, or vice versa, averaging or concatenation dilutes the informative cue with uninformative noise. When contextual labels are absent at test time, the network has no mechanism for reasoning about what the missing context implies.

To address this limitation, we propose CA-MoDE\footnote{To facilitate reproducibility, the PyTorch implementation of CA-MoDE is publicly available at~\url{https://github.com/dehshibi/CA-MoDE}.}, a context-aware mixture of domain experts that treats context integration as probabilistic evidence rather than feature augmentation. CA-MoDE uses dedicated scene and object experts to predict soft pseudo-labels conditioned on their respective domains. These predictions are compiled into emotion-context co-occurrence priors over the training set. These priors encode both the emotions supported by available contextual cues and the information carried by unavailable or weak contextual evidence. We also introduce a max-endorsement gating strategy that uses these priors to modulate the output of an independent emotion expert. By selecting the strongest contextual signal for each emotion dimension, the gate mitigates the dilution that occurs when informative and uninformative cues are averaged.

CA-MoDE is fully differentiable and trained end-to-end on the Body Language Database (BoLD)~\cite{luo2020arbee}. It achieves an Emotion Recognition Score (ERS) of 0.3269 using only single still images. This gain does not imply superiority of static over dynamic analysis. It does suggest that when contextual cues are explicitly modelled as structured probabilistic priors, they can partly compensate for the absence of temporal information, especially for categorical emotion recognition.

%=============SECTION II=================
\section{Related Work} \label{sec:survey}
Computational emotion recognition has historically been anchored in facial analysis. Early architectures exploited the Facial Action Coding System to define a tractable label space and achieved strong performance under controlled, frontal conditions~\cite{mollahosseini2016going}. Multi-task formulations later incorporated action unit detection to reduce the ambiguity of label-scarce settings~\cite{pons2020multitask}. These methods, however, failed to generalise in the wild. Psychological evidence shows that body language provides robust cues for emotion recognition in the wild, especially when faces are occluded or ambiguous~\cite{aviezer2012body}.

Interest in bodily expressions of emotion grew from these observations. The body carries broad postural and kinematic signals~\cite{dadamo2024soniweight}, but their interpretation depends strongly on the surrounding scene and the objects involved. Large-scale databases such as BoLD~\cite{luo2020arbee} made bodily expressions of emotion a tractable computational task. Luo et al.~\cite{luo2020arbee} introduced BoLD and a multi-task framework that jointly addresses pose estimation and emotion recognition, and the model must implicitly learn the influence of context on emotion.

Subsequent work on BoLD expanded the context integration pipeline while preserving this structure. Filntisis et~al.~\cite{filntisis2020emotion} combined skeleton sequences with multi-segment body features in a late-fusion ensemble. Pikoulis et~al.~\cite{9666957} incorporated scene attributes and optical flow over skeleton coordinates using a Spatial-Temporal Graph Convolutional Network. Li et~al.~\cite{li2021sequential} proposed SIB-Net, a hierarchical architecture that maps body crops against multi-scale background contexts through sequential interactive layers. A shared constraint of these models is also that contextual cues are not explicitly modelled, with their influence on the output distribution learned implicitly.

Recent advancements in Vision-Language Models (VLMs) have introduced an alternative paradigm for emotion recognition. Zhang et~al.~\cite{zhang2023emotionclip} applied contrastive learning on uncurated video captions with subject-aware attention masking to train EmotionCLIP, achieving a mean Average Precision (m$AP$) of 42.23\% and a mean ROC Area (m$RA$) of 81.36\% on BoLD. Similarly, Xenos et~al.~\cite{xenos2025vllm} demonstrated the effectiveness of semantic priors by using a pretrained VLM to generate scene descriptions, reporting an m$AP$ of 26.66\% and a classification accuracy of 93.08\%. The improvement demonstrates that structured semantic priors over scene content are more informative than raw feature concatenation. These results are promising, but they leave two issues unresolved. 

First, these VLM-based methods focus on specific classification subtasks of the BoLD benchmark rather than the full suite of continuous and discrete emotion predictions. Therefore, the unavailability of m$RA$ prevents computing the composite ERS, and, subsequently, a direct quantitative comparison is not applicable. Moreover, these methods depend on VLMs and provide no explicit mechanism for reasoning about what unavailable or weak contextual evidence implies.

Second, beyond the issue of metric comparability, the remaining gap is structural. Across conventional and recent VLM-based approaches, context is used to enrich inputs or align representations, yet the probabilistic link between context and emotion is rarely made explicit. When one contextual source provides a strong signal, and another is silent, feature alignment or fusion can attenuate the informative cue. When contextual evidence is missing or weak at test time, there is no explicit mechanism for reasoning about what that absence implies. CA-MoDE addresses this gap by computing emotion-context co-occurrence statistics from training data and using them as conditional distributions in the gating mechanism.

%=============SECTION III================
\section{Proposed Method}
\label{sec:method}

\begin{figure}[!htb]
    \centering
    \subfigure[]{
        \includegraphics[width=0.95\linewidth]{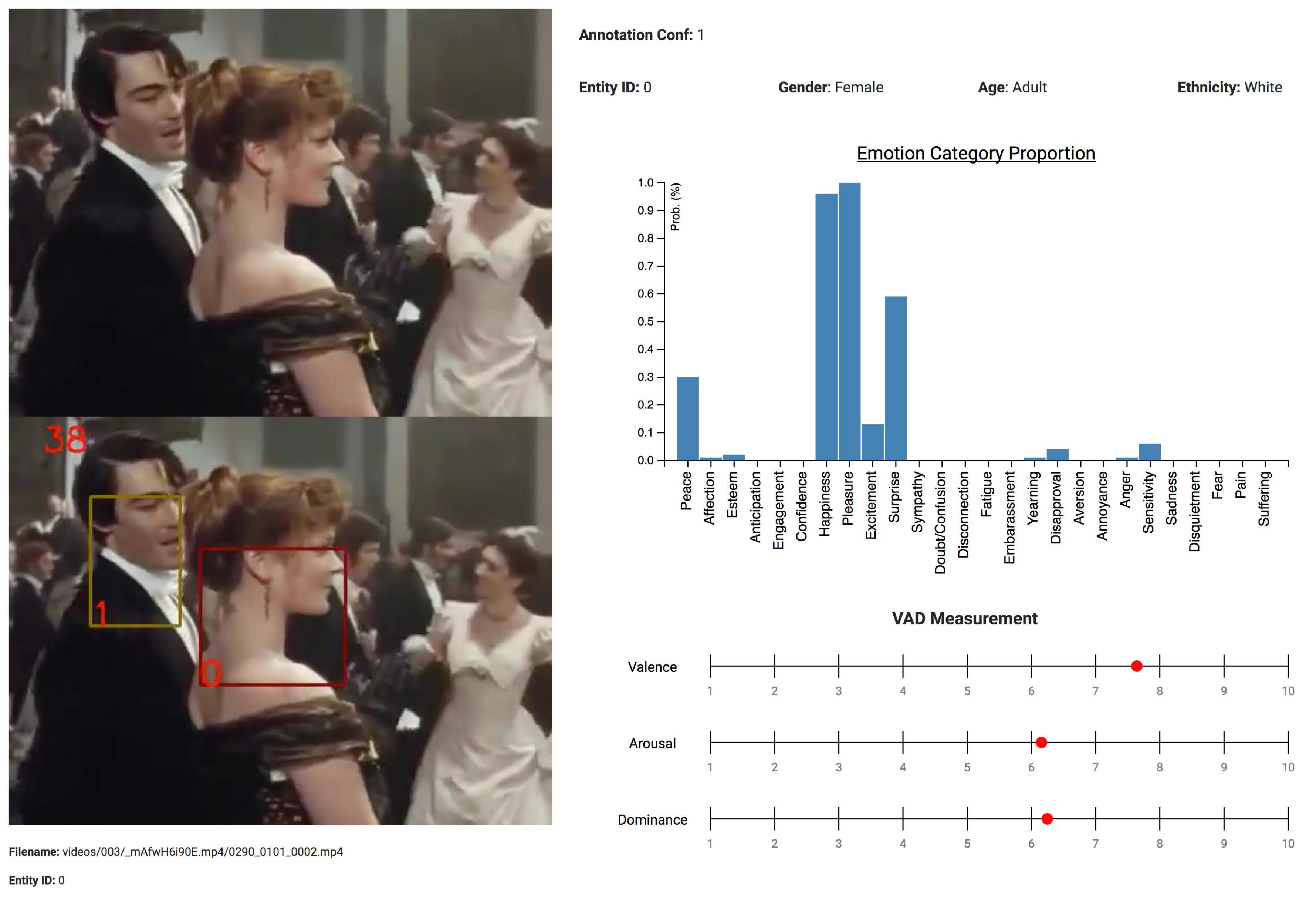}
        \label{fig:bold}}
    \subfigure[]{
        \includegraphics[width=0.577\linewidth]{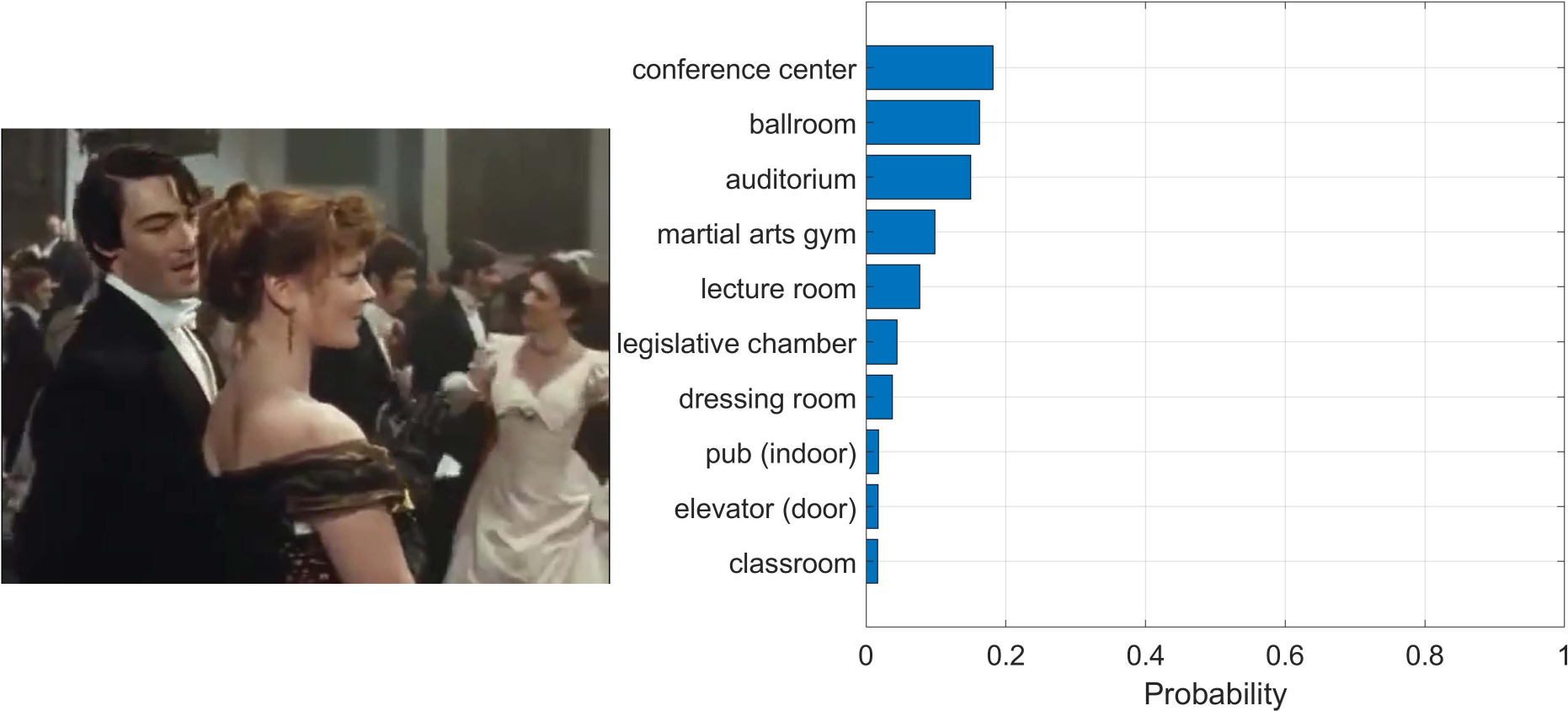}
        \label{fig:places}}
    \subfigure[]{
        \includegraphics[width=0.353\linewidth]{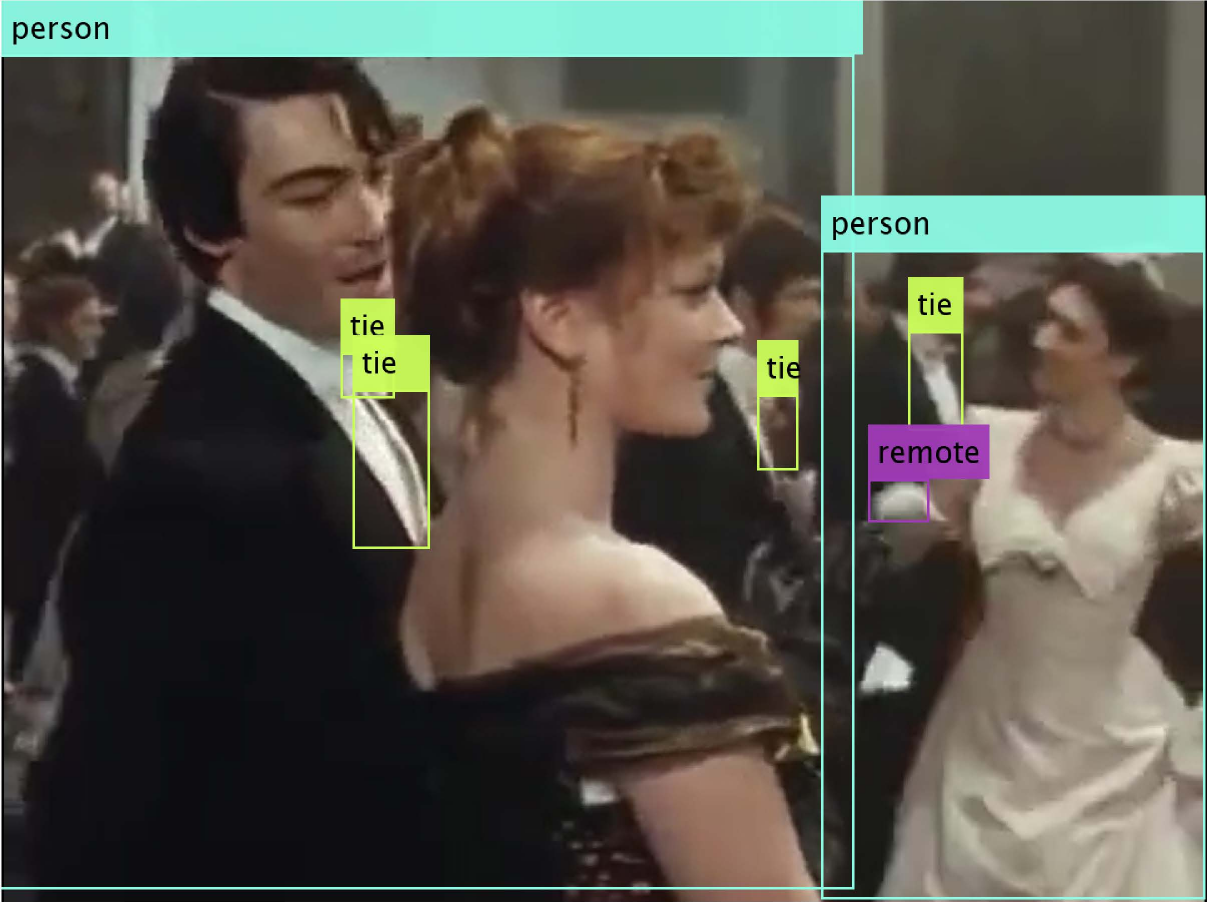}
        \label{fig:yolo}}
    \caption{(a) A sample from the BoLD database that mainly represents happiness. (b) Top-10 scene soft pseudo-labels obtained by applying Places-CNN, pre-trained on the Places2 database, to the input image. (c) Object soft pseudo-labels obtained by applying YOLO, pre-trained on the Microsoft COCO database, to the input image.}
    \label{fig:1}
\end{figure}

We formulate bodily emotion recognition as a multi-output regression problem. Given a dataset of $n$ images $\mathcal{D} = \{(x_{i}, y_{i})\}_{i=1}^{n}$, with $x_{i} \in \mathbb{R}^{\texttt{width} \times \texttt{height} \times \texttt{3}}$ and $y_{i} \in \mathbb{R}^{d_{y}},\ d_{y} = 29$, our goal is to train a deep neural network to predict the conditional expectation of the emotion labels given the input image. The target vector $y_{i}$ comprises 26 discrete emotions mapped to $[0,1]$ and three continuous affective variables (Valence, Arousal, Dominance; VAD) in $[1,10]$. Figure~\ref{fig:bold} shows emotional category proportion and VAD measurement for a scene. We optimise the network end-to-end by minimising the mean squared error between the predictions and the ground-truth annotations.

%---- PLACEMENT

%---- PLACEMENT
\begin{figure*}[!htbp]
    \centering
    \includegraphics[width=\linewidth]{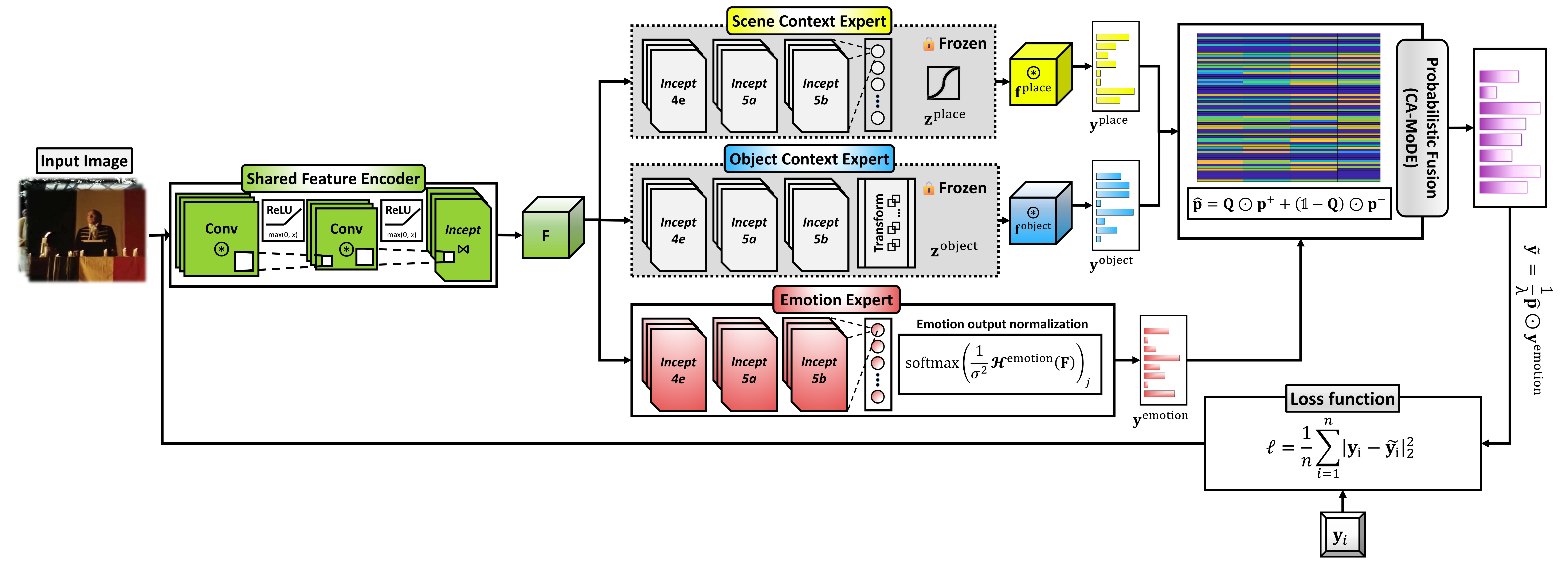}
    \caption{The proposed CA-MoDE architecture. The shared feature encoder is trainable, while the scene and object context expert backbones are frozen after initialisation from pre-trained models. During training, gradients flow only through the Emotion Expert and the learnable $1 \times 1$ convolution filters that project the contextual outputs into the fusion space.}
    \label{fig:architecture}
\end{figure*}

We assume that each image can be characterised by three complementary components, namely \textbf{emotion}, \textbf{place}, and \textbf{object}, each of which is handled by a dedicated domain expert. While emotion labels are explicitly provided in $\mathcal{D}$, contextual information (i.e., place and object) is not explicitly available in BoLD. To address this, we generate~\textbf{soft~pseudo-labels} for scene and object context (see Figs.~\ref{fig:places} and \ref{fig:yolo}) using domain experts pre-trained on the Places2~\cite{zhou2018places} and Microsoft COCO~\cite{lin2014microsoft} databases, respectively.

Mensink et~al.~\cite{mensink2014costa} showed that attribute co-occurrences observed during training tend to reappear at test time with high probability. Incorporating such co-occurrences helps reduce unwanted outliers introduced by domain integration and supports the use of $L_{2}$ as the regression loss for this task~\cite{yu2019unsupervised}. However, directly fine-tuning the contextual experts on BoLD is not possible because place and object labels are unavailable.

To address this problem, we propose a Context-Aware Mixture of Domain Experts (CA-MoDE) in which the probabilistic fusion mechanism channels contextual information into the emotion learning process using \textit{a priori} knowledge about the joint probability of emotions and both available and anticipated unavailable places and objects. We formulate a differentiable probabilistic fusion scheme, grounded in conditional probability identities, to combine the uncertain contextual information with the predicted image-based emotional states. Our fusion allows the network to discover correlations between context and emotion without additional post-processing or domain-specific regularisation. Figure~\ref{fig:architecture} illustrates the proposed architecture.

%-------------------------SECTION III.A-------------------------
\subsection{Context-Aware Mixture of Domain Experts Architecture}
\label{sec:network}

The proposed CA-MoDE comprises a shared feature encoder and three domain experts that process the extracted features in parallel. As illustrated in Fig.~\ref{fig:architecture}, the shared encoder provides a common feature representation to the scene context expert, the object context expert, and the emotion expert. The outputs of the contextual experts serve as soft pseudo-labels that feed into the probabilistic fusion mechanism described in Section~\ref{sec:fusion}.\newline

\noindent\textbf{(1) Shared Feature Encoder:} All domain experts share a common feature encoder $\mathcal{H}^{\text{base}}$ based on GoogLeNet~\cite{szegedy2015going}, where the output of the Inception 4d module serves as a shared feature representation. More formally, the shared feature map is computed as in Eq.~\eqref{eq:feat}.

\begin{equation}
    \label{eq:feat}
    \mathbf{F} = \mathcal{H}^{\text{base}}(\mathbf{X}),
    \quad \mathbf{F} \in \mathbb{R}^{W \times H \times D},
\end{equation}
where $W = 14$, $H = 14$, and $D = 528$ denote the width, height, and number of channels of the feature map, respectively. The tensor $\mathbf{F}$ is then fed to three domain experts specialised in scene context, object context, and bodily emotion cues.\newline

\noindent\textbf{(2) Scene Context Expert:} The scene context expert $\mathcal{H}^{\text{place}}$ is responsible for describing the environment in which the action takes place. It resembles the architecture of Places-CNN with a GoogLeNet backbone and includes the Inception 4e, 5a, and 5b modules, followed by global average pooling, dropout, a fully connected layer, and a softmax classifier. The scene expert produces a probability distribution over $365$ scene categories as formulated in Eq.~\eqref{eq:zplace}.

\begin{equation}
    \label{eq:zplace}
    \mathbf{z}^{\text{place}} = \mathcal{H}^{\text{place}}(\mathbf{F}), \quad \mathbf{z}^{\text{place}} \in \mathbb{R}^{1 \times 1 \times 365},
\end{equation}
where $\mathbf{z}^{\text{place}}$ contains soft pseudo-labels for scene context, i.e., predicted probabilities over Places2 categories. To harmonise the dimensionality of the contextual representations and prepare them for the probabilistic fusion described in Section~\ref{sec:fusion}, we apply a learnable $1 \times 1$ convolution followed by a bias term, as defined in Eq.~\eqref{eq:yplace}.

\begin{equation}
    \label{eq:yplace}
    \mathbf{y}^{\text{place}} = \mathbf{z}^{\text{place}} \circledast \mathbf{f}^{\text{place}} + \mathbf{b}^{\text{place}},
\end{equation}
where $\circledast$ denotes convolution, $\mathbf{f}^{\text{place}} \in \mathbb{R}^{1 \times 1 \times 365 \times \kappa}$ is a trainable filter bank, $\mathbf{b}^{\text{place}} \in \mathbb{R}^{\kappa}$ is a bias term, and $\kappa$ is the chosen latent dimensionality for contextual features. The weights of $\mathcal{H}^{\text{place}}$ are initialized from a Places2 pre-trained model and kept frozen during training on BoLD; only $\mathbf{f}^{\text{place}}$ and $\mathbf{b}^{\text{place}}$ are updated.\newline

\noindent\textbf{(3) Object Context Expert:} The object context expert $\mathcal{H}^{\text{object}}$ captures the set of objects present in the scene that may influence the perceived emotion. This expert is inspired by the YOLO detection architecture and consists of three groups of convolutional, ReLU, and batch normalisation layers. In the first group (Inception 4e), the filter size is set to $14 \times 14$ to match the spatial resolution of $\mathbf{F}$, whereas the second and third groups (Inception 5a and 5b) use $7 \times 7$ filters to facilitate the detection of smaller objects. The object expert produces a probability distribution over $80$ object categories as given in Eq.~\eqref{eq:zobject}.

\begin{equation}
    \label{eq:zobject}
    \mathbf{z}^{\text{object}} = \mathcal{H}^{\text{object}} (\mathbf{F}), \quad \mathbf{z}^{\text{object}} \in \mathbb{R}^{1 \times 1 \times 80},
\end{equation}
where $\mathbf{z}^{\text{object}}$ contains soft pseudo-labels for object context derived from a COCO pre-trained model. As in the scene expert, we project these probabilities into a $\kappa$-dimensional latent space using a $1 \times 1$ convolution, as defined in Eq.~\eqref{eq:yobject}.

\begin{equation}
    \label{eq:yobject}
    \mathbf{y}^{\text{object}} = \mathbf{z}^{\text{object}} \circledast \mathbf{f}^{\text{object}} + \mathbf{b}^{\text{object}},
\end{equation}
with $\mathbf{f}^{\text{object}} \in \mathbb{R}^{1 \times 1 \times 80 \times \kappa}$ and $\mathbf{b}^{\text{object}} \in \mathbb{R}^{\kappa}$. The parameters of $\mathcal{H}^{\text{object}}$ are initialised from a COCO pre-trained detector and kept frozen during training on BoLD; only $\mathbf{f}^{\text{object}}$ and $\mathbf{b}^{\text{object}}$ are updated. The vectors $\mathbf{y}^{\text{place}}$ and $\mathbf{y}^{\text{object}}$ constitute compact contextual representations that will be used in Section~\ref{sec:fusion} to estimate emotion-context co-occurrences and drive the probabilistic fusion mechanism.\newline

\noindent\textbf{(4) Emotion Expert:} The emotion expert $\mathcal{H}^{\text{emotion}}$ learns the mapping between the shared feature map $\mathbf{F}$ and the 29-dimensional emotion vector. It reuses the Inception 4e, 5a, and 5b modules, followed by global average pooling, dropout, and a fully connected layer. The output is formulated in Eq.~\eqref{eq:yemotion}.

\begin{equation}
    \label{eq:yemotion}
    \mathbf{y}^{\text{emotion}} = \mathcal{H}^{\text{emotion}}(\mathbf{F}), \quad \mathbf{y}^{\text{emotion}} \in \mathbb{R}^{1 \times d_{y}}.
\end{equation}

Although emotions can co-occur, we deliberately apply a precision-weighted softmax over the logits to sharpen the distribution toward the dominant emotion. This mechanism produces a discriminative prior for the gating module. Let $H_{j}(\mathbf{F})$ denote the $j$-th logit. The scaled class probabilities are given by Eq.~\eqref{eq:softmax}.

\begin{equation}
	\label{eq:softmax}
	p(y = j \mid \mathcal{H}^{\text{emotion}}(\mathbf{F}), \sigma) = \mathrm{softmax}\!\left( \frac{1}{\sigma^{2}} \mathcal{H}^{\text{emotion}}(\mathbf{F}) \right)_{j},
\end{equation}
where $\sigma > 0$ is a learnable scale parameter. Because the logits are multiplied by $1/\sigma^{2}$, a smaller $\sigma$ produces a sharper, more peaked distribution, while a larger $\sigma$ produces a softer distribution. The parameters of $\mathcal{H}^{\text{emotion}}$ are updated during training on BoLD using the multi-output regression loss defined in Section~\ref{sec:architectures}, unlike the contextual experts whose backbones are frozen.

%-------------------------SECTION III.B-------------------------
\subsection{Probabilistic Fusion of Contextual Experts} \label{sec:fusion}
We now derive the probabilistic fusion mechanism that modulates emotion predictions using contextual cues. To streamline notation, let $\mathbb{B}$ denote the emotion variable and $\mathbb{A}_{1}, \mathbb{A}_{2}$ denote scene and object contexts, respectively. These correspond to outputs from the emotion expert ($\mathcal{H}^{\text{emotion}}$), scene expert ($\mathcal{H}^{\text{place}}$), and object expert ($\mathcal{H}^{\text{object}}$).\newline

\noindent\textbf{Estimating Emotion-Context Co-occurrence:} We first compute empirical co-occurrence statistics between emotions and contextual cues over the training set. These statistics serve as priors for our probabilistic fusion mechanism. Let $N_{i}$ be the accumulated soft counts of the $i$-th emotion label over the training set, and $n$ the total number of training samples. We define the empirical prior probability of the $i$-th emotion by Eq.~\eqref{eq:3b01}.

\begin{equation}
    \label{eq:3b01}
    p_{i} = \mathrm{Pr}(\mathbb{B}_{i}) = \frac{N_{i}}{n}, \quad 1 \leq i \leq d_{y}.
\end{equation}

Similarly, let $N_{i,j}$ denote the weighted accumulated soft count between the $i$-th emotion and the $j$-th contextual expert, where $j \in \{1,2\}$ indexes the scene and object experts, respectively. For each sample, we summarise the output of expert $j$ by a scalar confidence score $s_{k,j} \in [0,1]$. This score, defined as the maximum predicted category probability, is then used to compute $N_{i,j} = \sum_{k=1}^{n} y_{k, i} \times s_{k, j}$. The corresponding joint probability estimate is calculated using Eq.~\eqref{eq:3b02}.

\begin{equation}
    \label{eq:3b02}
    \mathcal{C}_{i,j} = \mathrm{Pr}(\mathbb{B}_{i} \cap \mathbb{A}_{j}) = \frac{N_{i,j}}{n}.
\end{equation}

Using these priors, we can approximate the conditional probability of an emotion given a contextual event using Eq.~\eqref{eq:3b03}.

\begin{equation}
    \label{eq:3b03}
    \mathcal{P}^{+}_{i,j} = \mathrm{Pr}(\mathbb{B}_{i}|\mathbb{A}_{j}) = \frac{\mathrm{Pr}(\mathbb{B}_{i} \cap \mathbb{A}_{j})} {\mathrm{Pr}(\mathbb{A}_{j})} = \frac{\mathcal{C}_{i,j}}{q_{j}},
\end{equation}
where $q_{j} = \mathrm{Pr}(\mathbb{A}_{j})$ is the prior probability of the $j$-th contextual event. It is estimated from training data by aggregating the predicted soft pseudo-labels over the dataset. We collect these conditional probabilities into $\mathcal{P}^{+} \in \mathbb{R}^{d_{y} \times 2}$ as in Eq.~\eqref{eq:3b04}.

\begin{equation}
    \label{eq:3b04}
    \mathcal{P}^{+} =
    \begin{bmatrix}
        \mathrm{Pr}(\mathbb{B}_{1}|\mathbb{A}_{1}) &
        \mathrm{Pr}(\mathbb{B}_{2}|\mathbb{A}_{1}) &
        \cdots &
        \mathrm{Pr}(\mathbb{B}_{d_{y}}|\mathbb{A}_{1}) \\
        \mathrm{Pr}(\mathbb{B}_{1}|\mathbb{A}_{2}) &
        \mathrm{Pr}(\mathbb{B}_{2}|\mathbb{A}_{2}) &
        \cdots &
        \mathrm{Pr}(\mathbb{B}_{d_{y}}|\mathbb{A}_{2})
    \end{bmatrix}^{\top},
\end{equation}
where the first and second columns correspond to scene and object context, respectively.\newline

\noindent\textbf{Modelling Anticipated Unavailable Context:} While $\mathcal{P}^{+}$ captures what available context makes likely, the absence or low confidence of a contextual cue can also be informative. We refer to this signal as anticipated unavailable context and model it as $\mathcal{P}_{i,j}^{-} = \mathrm{Pr} (\mathbb{B}_{i}|\neg\mathbb{A}_{j})$. Using standard conditional probability theory, we express $\mathrm{Pr}(\mathbb{B}_{i}|\neg\mathbb{A}_{j})$ in terms of marginal and joint probabilities using Eq.~\eqref{eq:3b05}.

\begin{equation}
    \label{eq:3b05}
    \mathrm{Pr}(\mathbb{B}_{i}|\neg\mathbb{A}_{j}) = \frac{\mathrm{Pr}(\mathbb{B}_{i} \cap \neg\mathbb{A}_{j})} {\mathrm{Pr}(\neg\mathbb{A}_{j})} = \frac{\mathrm{Pr} (\mathbb{B}_{i}) - \mathrm{Pr}(\mathbb{B}_{i} \cap \mathbb{A}_{j})} {1 - q_{j}}.
\end{equation}

Substituting our empirical estimates leads to $\mathcal{P}_{i,j}^{-} = \frac{p_{i} - \mathcal{C}_{i,j}}{1 - q_{j}}$, where we collect these values into a matrix $\mathcal{P}^{-} \in \mathbb{R}^{d_y \times 2}$. This formulation allows us to reason not only about \textit{what is present} in the scene, but also about \textit{what is expected to be absent}, both of which can significantly influence perceived emotions.\newline

\noindent\textbf{Probabilistic Fusion of of Contextual Priors:} Given $\mathcal{P}^{+}$, $\mathcal{P}^{-}$, we first aggregate the contextual contributions for each emotion using Eq.~\eqref{eq:3b06b}.
\begin{equation}
    \label{eq:3b06b}
    p^{+}_{i} = \max_{j \in \{1,2\}} \mathcal{P}^{+}_{i,j},
    \quad
    p^{-}_{i} = \max_{j \in \{1,2\}} \mathcal{P}^{-}_{i,j},
\end{equation}
and stack these into vectors $\mathbf{p}^{+}, \mathbf{p}^{-} \in \mathbb{R}^{1 \times d_{y}}$. The max operation implements a winner-take-all selection for each emotion dimension, mitigating dilution of strong single-expert signals. These aggregated vectors are then combined by a gate that balances available contextual support against the missing-context prior.\newline

\noindent\textbf{Gating Weights from Contextual Support:} To combine $\mathbf{p}^{+}$ and $\mathbf{p}^{-}$, we define a gating vector $\mathbf{Q} \in [0,1]^{1 \times d_{y}}$ that, for each emotion dimension $i$, determines how much weight to assign to the available-context prior versus the missing-context prior. Let $Q_{i}$ denote the $i$-th entry of $\mathbf{Q}$. Intuitively, if the aggregated available endorsement $p^{+}_{i}$ is strong, then $Q_{i}$ should be high, giving more weight to available context. Conversely, if $p^{+}_{i}$ is weak, $Q_{i}$ should be low, and the probabilistic fusion should rely more on the anticipated unavailable context $p^{-}_{i}$. We define $Q_{i}$ as in Eq.~\eqref{eq:3b06}.

\begin{equation}
    \label{eq:3b06}
    Q_{i} = \mathrm{sigmoid} \Big(\alpha \big(p^{+}_{i}-\tau\big) \Big), \quad 1 \leq i \leq d_{y}.
\end{equation}

The sigmoid gate maps the strongest available contextual endorsement per emotion, $p^{+}_{i}$, to a soft weight in $(0,1)$, where $\tau$ controls the threshold above which contextual support is considered strong and $\alpha$ controls the sharpness of the transition. We fix $\tau = 0.5$ as the threshold and $\alpha = 10$ to control gate sharpness across all experiments. These values were selected via validation set grid search over $\tau \in \{0.3, 0.4, 0.5, 0.6, 0.7, 0.8, 0.9\}$ and $\alpha \in \{5, 10, 15, 20\}$. 

Under this gate, higher $Q_i$ values indicate stronger contextual endorsement and make the fused estimate rely more on the available-context term $p_i^{+}$, whereas lower values shift it towards the anticipated unavailable-context term $p_i^{-}$. Empirically, this simple interpolation provides a favourable balance between simplicity and robustness, especially in the presence of noisy soft pseudo-labels. We then compute a convex combination of the two contextual priors for each emotion using Eq.~\eqref{eq:3b07}.

\begin{equation}
    \label{eq:3b07}
    \hat{p}_{i} = Q_{i} p^{+}_{i} + (1 - Q_{i}) p_{i}^{-}, \quad 1 \leq i \leq d_{y}.
\end{equation}

In vector form, $\hat{\mathbf{p}} = \mathbf{Q} \odot \mathbf{p}^{+} + (\mathbb{1} - \mathbf{Q}) \odot \mathbf{p}^{-}$, where $\odot$ denotes element-wise multiplication and $\mathbb{1}$ is the all-ones vector of size $1 \times d_{y}$. The fused vector $\hat{\mathbf{p}}$ encodes, for each emotion, the combined influence of available and missing contextual evidence, weighted by the strength of contextual endorsement. It is grounded in conditional probability identities but designed as a robust, differentiable heuristic rather than a strict Bayesian posterior.\newline

\noindent\textbf{Emotion Prediction and Training Objective:} Finally, the fused contextual vector $\hat{\mathbf{p}}$ is used to modulate the output of the emotion expert, reinforcing emotion predictions that are consistent with the contextual priors while suppressing those that are not. The predicted emotion vector is defined using Eq.~\eqref{eq:pred}.

\begin{equation}
    \label{eq:pred}
    \tilde{\mathbf{y}} = \frac{1}{\lambda}\,\hat{\mathbf{p}} \odot \mathbf{y}^{\text{emotion}},
\end{equation}
where $\lambda > 0$ is a scalar hyperparameter selected on the validation set to control the scale of the predictions. Intuitively, entries of $\hat{\mathbf{p}}$ close to one amplify the corresponding emotion predictions, while entries close to zero suppress predictions that neither the available nor the anticipated unavailable context supports.

Given ground-truth labels $\{\mathbf{y}_{i}\}_{i=1}^{n}$ from BoLD, we train the complete network end-to-end by minimizing the mean squared error using Eq.~\eqref{eq:loss}.

\begin{equation}
    \label{eq:loss}
    \ell = \frac{1}{n} \sum_{i=1}^{n} \left\| \mathbf{y}_{i} -  \tilde{\mathbf{y}}_{i} \right\|_{2}^{2}.
\end{equation}

Following~\cite{luo2020arbee}, we obtain classification metrics (m$AP$, m$RA$) by thresholding $\tilde{\mathbf{y}}_{1:26}$ at $0.5$. We directly evaluate the regression metric (mR$^{2}$) on $\tilde{\mathbf{y}}_{27:29}$. The gradients propagate through the probabilistic fusion mechanism and the emotion expert, allowing the network to adapt the mapping from shared features and contextual signals to the final emotion representation.

%=============SECTION IV=================
\section{Architecture Details} \label{sec:architectures}

Experiments were performed on the NVIDIA GeForce RTX 5090 GPU (24 GB). All models were implemented in PyTorch and trained from scratch or fine-tuned as described below. All hyperparameter selection is performed on the validation set, with final evaluation on the held-out test set.\newline

\noindent\textbf{Scene Context Expert:} In $\mathcal{H}^{\text{place}}$, we initialise the weights and parameters of all layers to those of Places-CNN, which was pre-trained on the Places2 database~\cite{zhou2018places}. We set the learning rates of all its layers to zero to ensure that the output distribution of $\mathcal{H}^{\text{place}}$ remains consistent with the scene soft pseudo-labels used to compute $\mathcal{P}^{+}$ and $\mathcal{P}^{-}$ during training.\newline

\noindent\textbf{Object Context Expert:} In $\mathcal{H}^{\text{object}}$, we replace the pooling layers with strided convolutions of strides $16 \times 16$ and $32 \times 32$, respectively, to preserve spatial information while reducing the feature map resolution. We pre-train $\mathcal{H}^{\text{base}} + \mathcal{H}^{\text{object}}$ jointly on the Microsoft COCO database~\cite{lin2014microsoft} using a YOLO-style configuration with 7 anchor boxes and a non-maximal suppression threshold of 0.4. We then transfer the weights of $\mathcal{H}^{\text{object}}$ to CA-MoDE and set the learning rates to zero, ensuring the output distribution remains consistent with the object soft pseudo-labels used to compute $\mathcal{P}^{+}$ and $\mathcal{P}^{-}$.\newline

%--PLACEMENT
\begin{figure*}[!htbp]
    \centering
    \subfigure[]{\includegraphics[width=0.32\linewidth]{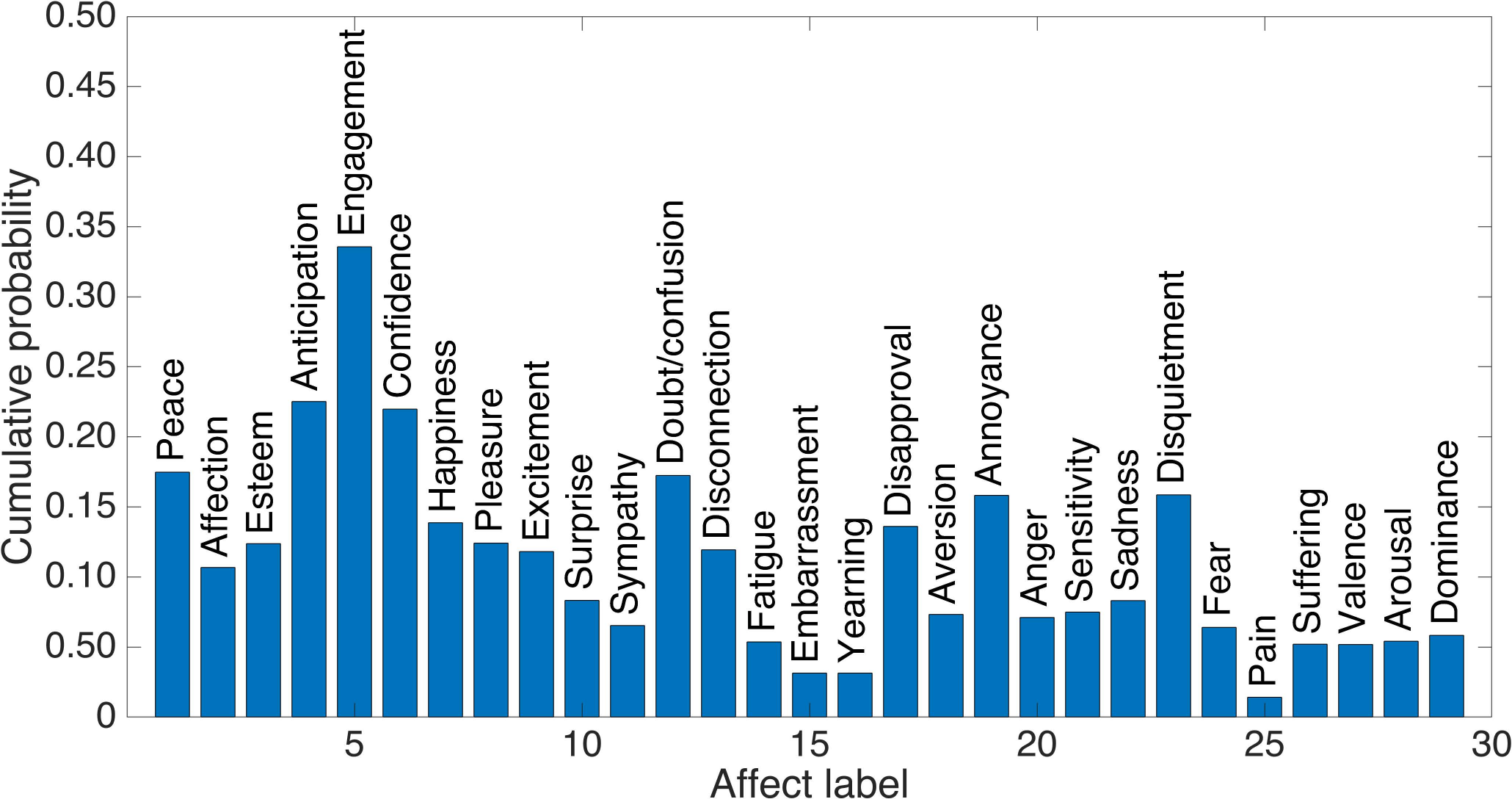} \label{fig:partition}}
    \subfigure[]{\includegraphics[width=0.32\linewidth]{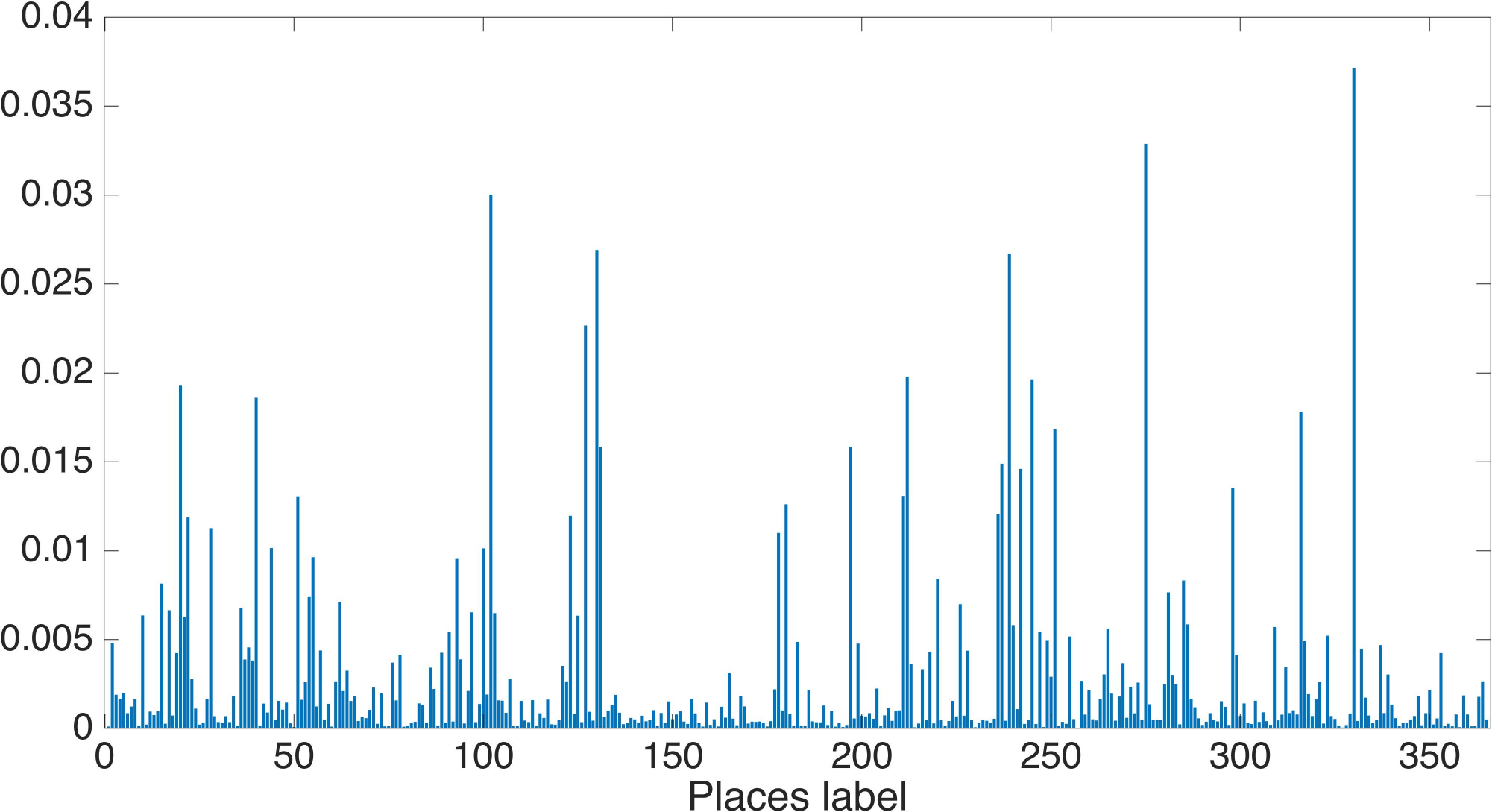} \label{fig:Pdistribution}}
    \subfigure[]{\includegraphics[width=0.32\linewidth]{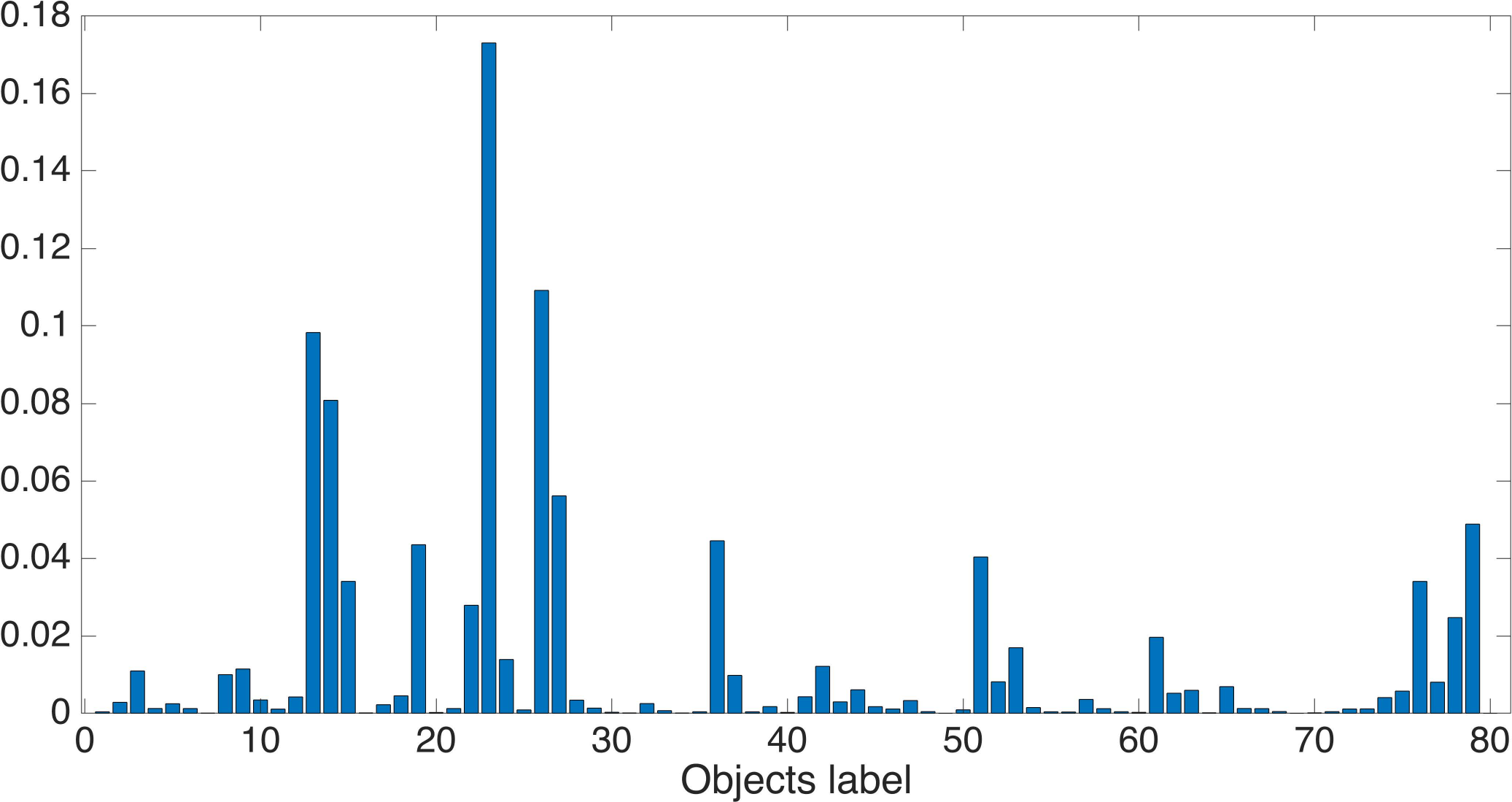}\label{fig:Odistribution}}
    \caption{Cumulative probability of (a) emotion labels in the BoLD database, (b) scene soft pseudo-labels obtained by applying the scene context expert to the BoLD training set, and (c) object soft pseudo-labels obtained by applying the object context expert to the BoLD training set.}
\end{figure*}

\noindent\textbf{Emotion Expert:} In $\mathcal{H}^{\text{emotion}}$, we train the network using stochastic gradient descent with a momentum of 0.9. The initial learning rate is set to $10^{-2}$ and is decreased by a factor of 0.1 every 45 epochs, with a maximum of 90 training epochs. We use a mini-batch of 8 per iteration, and the training data is shuffled before each epoch. The training parameters for the shared feature encoder $\mathcal{H}^{\text{base}}$ are set to the same values.\newline

\noindent\textbf{Hyperparameter selection:} The CA-MoDE architecture has two tuned hyperparameters, $\kappa$ and $\lambda$. The parameter $\kappa$ controls the output dimensionality of the learnable filter banks in the context experts. The parameter $\lambda$ controls the scale of the final prediction $\tilde{\mathbf{y}}$. To determine the appropriate range for $\kappa$, we apply the pre-trained scene and object context experts to the training set.

The context experts map input images $x_{i}$ to their respective class. For the scene context expert, the softmax output, $z_{i}^{\text{place}}$, is a 365-dimensional vector containing the probability of each scene category. For the object context expert, the softmax output, $z_{i}^{\text{object}}$, has 80 entries, one per COCO category. The list of detected objects per image may contain multiple instances of the same category with different confidence scores. To produce a fixed-length representation, we assign the normalised cumulative confidence score of all detected instances to the corresponding entry and set the remaining entries to zero.

The maximum value of $\kappa$ is bounded by $\min(365, 80) = 80$. To identify the minimum meaningful value of $\kappa$, we compute the normalised average soft pseudo-label vectors over the training set for both scene and object contexts, whose distributions are shown in Figs.~\ref{fig:Pdistribution} and~\ref{fig:Odistribution}, respectively. Applying a threshold of 0.01, we find that 27 place and 14 object categories exceed this threshold, demonstrating that the vast majority of entries in both vectors are negligible. We therefore evaluate $\kappa$ from 14 to 80 in steps of 6.

\begin{figure}[!htbp]
    \centering
    \includegraphics[width=0.80\linewidth]{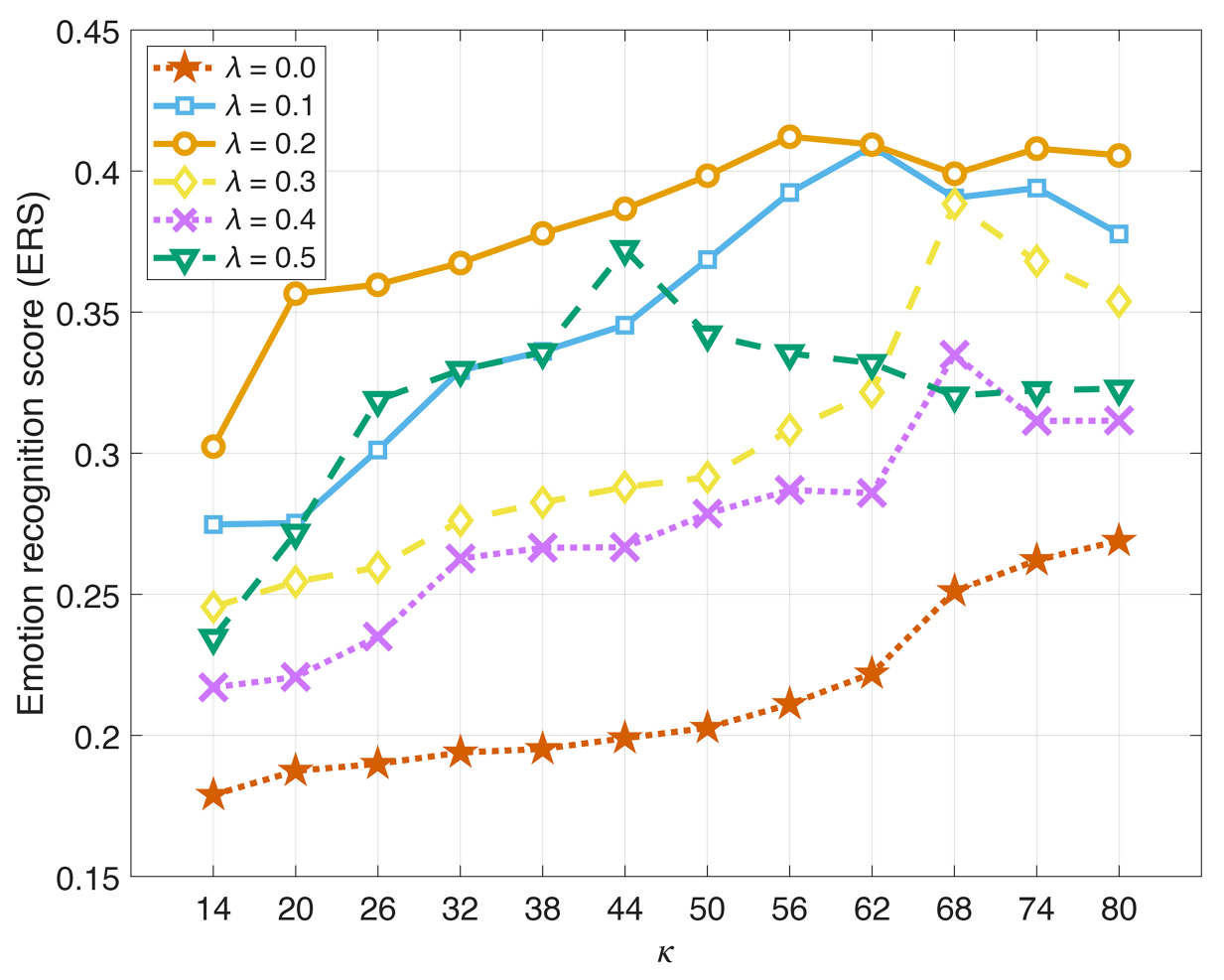}
    \caption{Emotion Recognition Score (ERS) on the validation set for CA-MoDE as a function of $(\kappa, \lambda)$. The best trade-off is achieved at $(\kappa, \lambda) = (56, 0.2)$ with an ERS of 0.4122.}
    \label{fig:ers_kappa_lambda}
\end{figure}

Similarly, we evaluate the impact of $\lambda$ by varying it from 0 to 0.5 in steps of 0.1. The analysis of the trade-offs across all metrics on the validation set shows that $(\kappa, \lambda) = (56, 0.2)$ allows CA-MoDE to achieve an ERS of 0.4122 (see Fig.~\ref{fig:ers_kappa_lambda}). We use these values in all subsequent experiments.

%---PLACEMENT

From Fig.~\ref{fig:ers_kappa_lambda} and an analysis of prediction errors, we observe that larger values of $(\kappa, \lambda)$ not only increase computational complexity but also make the prediction $\tilde{\mathbf{y}}$ more sensitive to the contextual modulation terms. This increased sensitivity reduces robustness when contextual soft pseudo-labels are noisy or uninformative.

%=============SECTION V==================
\section{Experiments} \label{sec:experiment}

%-------------------------SECTION V.A-------------------------
\subsection{Body Language Database (BoLD)} \label{sec:bold}

Body Language Dataset (BoLD)~\cite{luo2020arbee} is constructed from video clips drawn from the AVA database~\cite{gu2018ava}, a collection of YouTube movie clips comprising 9,876 clips and 13,239 isolated human instances. BoLD annotations\footnote{The discrete categorical labels include Peace, Affection, Esteem, Anticipation, Engagement, Confidence, Happiness, Pleasure, Excitement, Surprise, Sympathy, Doubt/Confusion, Disconnection, Fatigue, Embarrassment, Yearning, Disapproval, Aversion, Annoyance, Anger, Sensitivity, Sadness, Disquietment, Fear, Pain, and Suffering.}\textsuperscript{,}\footnote{The continuous affective variables (VAD) are Valence (V), Arousal (A), and Dominance (D).} were obtained through crowdsourcing on Amazon Mechanical Turk to establish multi-annotator consensus for highly subjective emotional states.  

%-------------------------SECTION V.B-------------------------
\subsection{Evaluation Metrics and Experimental Protocol}

We use the mean $R^{2}$ score to evaluate the regression component of the proposed model. For the $k$-th continuous emotion dimension, $R^{2}$ measures the ratio of variance explained by the model to the total variance of the ground truth, as defined in Eq.~\eqref{eq:r2}.

\begin{equation}
    \label{eq:r2}
    R^{2}(y^{(k)},\tilde{y}^{(k)}) = 1 -
    \frac{\displaystyle\sum_{i=1}^{n}
          \bigl(y_{i}^{(k)}-\tilde{y}_{i}^{(k)}\bigr)^{2}}
         {\displaystyle\sum_{i=1}^{n}
          \bigl(y_{i}^{(k)}-\varepsilon^{(k)}\bigr)^{2}},
    \quad
    \varepsilon^{(k)} = \frac{1}{n}\sum_{i=1}^{n}y_{i}^{(k)},
\end{equation}
where $y_{i}^{(k)}$ and $\tilde{y}_{i}^{(k)}$ are the ground-truth and predicted values of the $i$-th sample for dimension $k$, respectively. The mean coefficient of determination is then $mR^{2} = \frac{1}{3}\sum_{k=1}^{3} R^{2}(y^{(k)},\tilde{y}^{(k)})$, averaged over the VAD dimensions.

%--PLACEMENT
\begin{figure*}[!htbp]
    \centering
    \includegraphics[width=0.95\linewidth]{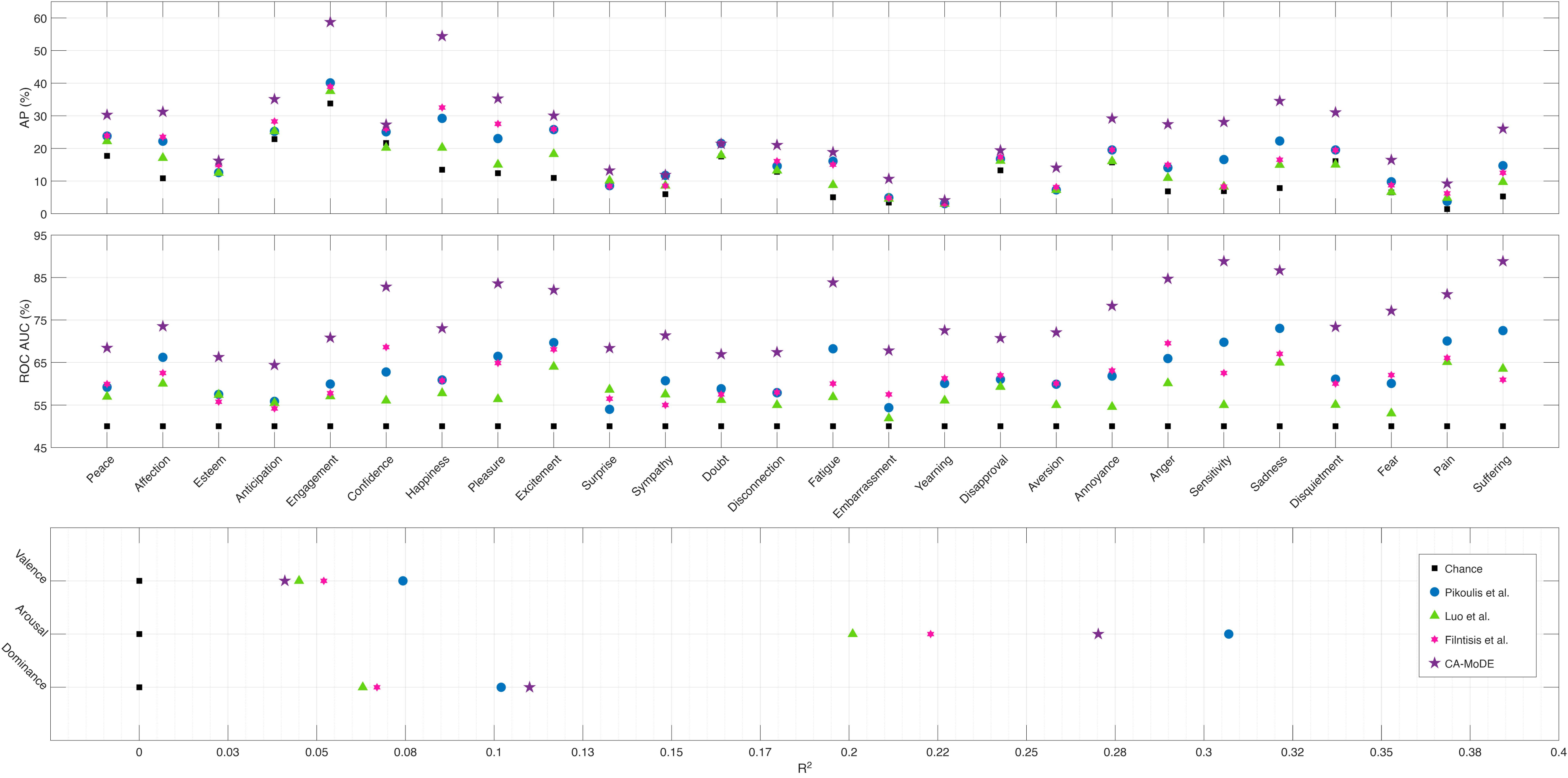}
    \caption{Per-emotion metric comparisons on the BoLD test set. Classification performance for discrete emotions is reported based on the m$AP$ in the first row and m$RA$ in the second row. Regression performance for the VAD is reported based on the m$R^{2}$ score in the third row.}
    \label{fig:metrics}
\end{figure*}

Since we formulate the AIBEE task as a multi-output regression problem, classification metrics for the 26 discrete emotions are obtained by thresholding the continuous predictions $\tilde{\mathbf{y}}_{1:26}$ at 0.5. For each of the 26 emotion dimensions ($k \in \{1,\ldots,26\}$), we compute Average Precision $AP^{(k)}$ and Receiver Operating Characteristic Area $RA^{(k)}$, and report their means. For ease of comparison across methods, the Emotion Recognition Score (ERS) aggregates regression and classification performance into a single scalar using Eq.~\eqref{eq:ers}.

\begin{equation}
    \label{eq:ers}
    \mathrm{ERS} = \frac{1}{2}\left( \mathrm{m}R^{2} + \frac{1}{2}(\mathrm{m}AP + \mathrm{m}RA) \right).
\end{equation}

We use the standard BoLD evaluation protocol, partitioning the dataset into training (60\%), validation (10\%), and test sets (30\%). The partitioning was performed at the clip level to ensure that no two frames from the same video are assigned to different sets, while preserving the BoLD class distribution within each split. All hyperparameter selection is performed on the validation set, and the final evaluation is reported exclusively on the held-out test set.

%-------------------------SECTION V.C-------------------------
\subsection{Experimental results} \label{sec:results}
Table~\ref{tbl:1} reports the performance of CA-MoDE alongside competitive methods on the BoLD test set. Following the evaluation protocol of~\cite{luo2020arbee}, we include a random method based on priors (Chance) as a lower-bound reference. The comparison includes the BoLD baseline~\cite{luo2020arbee}, the BEEU challenge winner~\cite{filntisis2020emotion}, and the top-ranked method on the BoLD leaderboard~\cite{9666957}, all of which leverage temporal information across video frames through optical flow, skeleton sequences, or multi-segment sampling, and all of which report the three metrics required to compute ERS.

\begin{table}[!htbp]
    \centering
    \caption{Performance on the BoLD test set.}
    \label{tbl:1}
    \resizebox{\linewidth}{!}{%
    \begin{tabular}{lllll}
        \hline
        \multirow{2}{*}{\textbf{Method}} & \multirow{2}{*}{\textbf{ERS}} &  \textbf{Regression} &  \multicolumn{2}{c}{\textbf{Classification}} \\  \cline{3-5} & & \boldmath m$R^{2}\uparrow$ & \boldmath m$AP\uparrow$ & \boldmath m$RA\uparrow$ \\ \hline
        Chance & 0.1513 & 0 & 0.1055 & 0.5000 \\
        Luo et al.~\cite{luo2020arbee} & 0.2531 & 0.1030 & 0.1714 & 0.6352 \\
        Filntisis et al.~\cite{filntisis2020emotion} & 0.2623 & 0.1141 & 0.1796 & 0.6416 \\
        Pikoulis et al.~\cite{9666957} & 0.3058 & \textbf{0.1609} & 0.2187 & 0.6829 \\
        \textbf{CA-MoDE} & \textbf{0.3269} & 0.1404 & \textbf{0.2938} & \textbf{0.7333} \\ \hline
    \end{tabular}}
\end{table}

What makes this result noteworthy is that all three competing methods exploit temporal information across video frames using optical flow, multi-segment sampling, or skeleton sequences. In contrast, CA-MoDE operates on single still images and relies exclusively on scene and object context for its reasoning. The fact that a still-image method can match or even surpass video-based approaches on overall ERS and, by a clear margin, on classification metrics, highlights the contribution of spatial contextual cues relative to temporal dynamics.

The lower m$R^{2}$ reflects a deliberate architectural trade-off. CA-MoDE conditions emotion predictions on discrete contextual evidence from scene and object experts. This conditioning is more beneficial for classifying categorical emotions than for regressing continuous dimensions. The resulting gains in m$AP$ and m$RA$ over the previous best are sufficiently large that the aggregate ERS improves despite the regression deficit.

Figure~\ref{fig:metrics} provides per-emotion comparisons across all methods. The emotion categories \texttt{Engagement}, \texttt{Happiness}, \texttt{Pleasure}, \texttt{Anticipation}, \texttt{Sadness} constitute the top-5 most accurately predicted emotions. This is particularly notable given that the BoLD database is biased towards \texttt{Engagement}, \texttt{Anticipation}, \texttt{Confidence}, \texttt{Peace}, \texttt{Doubt/Confusion} (see Fig.~\ref{fig:partition}). The top-5 predicted categories only partially overlap with the five most frequent categories. This divergence suggests that the probabilistic fusion mechanism in CA-MoDE attenuates the effect of label imbalance by conditioning predictions on contextual evidence rather than on marginal label frequency alone. In the ablation study, we demonstrate that explicitly modelling both available and anticipated unavailable context directly contributes to this behaviour.

In the assessment of continuous emotions, all methods demonstrate stronger regression performance for arousal than for valence and dominance, consistent with the subjective human evaluation gap reported in~\cite{luo2020arbee}. For CA-MoDE specifically, this asymmetry reflects the unequal availability of static cues. Arousal correlates with postural and scene-level features that a single frame can convey, whereas valence and dominance require finer disambiguation that $\mathcal{P}^{+}$ and $\mathcal{P}^{-}$ only partially provide. The modest $mR^{2}$ values across all methods confirm that the full VAD regression task remains an open challenge beyond the reach of spatial contextual reasoning alone.

%---- PLACEMENT
\begin{table*}[!htbp]
    \centering
    \caption{Impact of the underlying network architecture on AIBEE performance.}
    \label{tbl:backbone}
    \resizebox{0.76\linewidth}{!}{%
        \begin{tabular}{lccccc}
            \hline
            \multirow{2}{*}{\textbf{Architecture Family}} & \multicolumn{1}{l}{\multirow{2}{*}{\textbf{Trainable Parameters (M)}}} & \multirow{2}{*}{\textbf{ERS}} & \textbf{Regression} & \multicolumn{2}{c}{\textbf{Classification}} \\ \cline{4-6} 
             & \multicolumn{1}{l}{} &  & \boldmath m$R^{2}$ & \boldmath m$AP$ & \boldmath m$RA$ \\ \hline
             \multicolumn{6}{l}{\cellcolor[HTML]{EFEFEF}Ablation} \\ \hline
            ResNet-18~\cite{he2016deep} & 10.51 & 0.3206 & \textbf{0.1398} & 0.2815 & 0.7216 \\
            EVA-02~\cite{fang2024eva} & 273.15 & \textbf{0.3272} & 0.1392 & \textbf{0.3067} & 0.7238 \\ \hline
            \multicolumn{6}{l}{\cellcolor[HTML]{EFEFEF}Proposed architecture} \\ \hline
            CA-MoDE (GoogLeNet-derived) & \textbf{6.11} & 0.3269 & \textbf{0.1404} & 0.2938 & \textbf{0.7333} \\ \hline
        \end{tabular}%
    }
\end{table*}

%-------------------------SECTION V.D-------------------------
\subsection{Ablation study}
\label{sec:ablation}
In the ablation study, we conducted six sets of experiments to quantify the contribution of each component of CA-MoDE:\newline

\noindent\textbf{(1) Contribution of Underlying Network Architecture:} To assess whether CA-MoDE's performance depends on a specific network family, we replace the GoogLeNet-derived modules across the shared encoder and all three domain experts with ResNet-18~\cite{he2016deep} and EVA-02~\cite{fang2024eva} equivalents. Table~\ref{tbl:backbone} reports the trainable parameter count for each configuration, together with ERS, regression, and classification performance. EVA-02 achieves marginally higher ERS and m$AP$, consistent with the benefits of large-scale transformer pre-training, but at roughly 45 times the trainable parameter count of the proposed architecture. The proposed architecture matches this performance while outperforming both alternatives on regression and m$RA$, using substantially fewer trainable parameters. These results indicate that CA-MoDE's gains are attributable to the proposed probabilistic fusion mechanism rather than to the capacity of the underlying network architecture, and that the formulation does not require large-capacity networks to achieve competitive performance.\newline

\noindent\textbf{(2) Contribution of Pre-trained Model Weight:} In this experiment, instead of assigning random weights to the $\mathcal{H}^{base} + \mathcal{H}^{emotion}$, we evaluated GoogLeNet weights pre-trained on ImageNet, Places2, and Microsoft COCO. We reduced the training schedule to 45 epochs while retaining all other parameters as described in Section~\ref{sec:architectures}. The results in Table~\ref{tbl:2} show that random initialisation leads to marginally better performance across all metrics, suggesting that BoLD's visual statistics are sufficiently distinct from those of general-purpose databases.

\begin{table}[!htbp]
    \centering
    \caption{Impact of initial weights on AIBEE performance.}
    \label{tbl:2}
    \resizebox{0.95\linewidth}{!}{%
        \begin{tabular}{lllll}
            \hline
            \multirow{2}{*}{\textbf{Initial weight}} & \multirow{2}{*}{\textbf{ERS}} & \textbf{Regression} & \multicolumn{2}{c}{\textbf{Classification}} \\ \cline{3-5} 
             &  & \boldmath m$R^{2}$ & \boldmath m$AP$ & \boldmath m$RA$ \\ \hline
            \multicolumn{5}{l}{\cellcolor[HTML]{EFEFEF}Ablation} \\ \hline
            Random & \textbf{0.3041} & 0.1141 & \textbf{0.2901} & \textbf{0.6982} \\
            ImageNet & 0.2990 & \textbf{0.1218} & 0.2889 & 0.6636 \\
            Places2 & 0.2956 & 0.1103 & 0.2702 & 0.6918 \\
            Microsoft COCO & 0.2858 & 0.1097 & 0.2836 & 0.6402 \\ \hline
            \multicolumn{5}{l}{\cellcolor[HTML]{EFEFEF}Proposed architecture} \\ \hline
            CA-MoDE & 0.3269 & 0.1404 & 0.2938 & 0.7333 \\ \hline
        \end{tabular}
    }
\end{table}
    
\noindent\textbf{(3) Contribution of \mbox{\boldmath $\mathcal{H}^{\text{place}}$} and \mbox{\boldmath $\mathcal{H}^{\text{object}}$}:} Having confirmed that the results are not driven by initialisation, in this experiment, we analysed which of the two context experts contributes more. We independently ablated the scene context expert ($\mathcal{H} \circleddash \mathcal{H}^{\text{place}}$) and the object context expert ($\mathcal{H} \circleddash \mathcal{H}^{\text{object}}$) by removing each in turn and retraining on BoLD. All parameters were retained as described in Section~\ref{sec:architectures}, except that $\kappa$ was set to 100 when only the scene context expert was active and to 40 when only the object context expert was active. The results are presented in Table~\ref{tbl:3}.

\begin{table}[!htbp]
    \centering
    \caption{Ablation results for removing scene and object context experts on BoLD.}
    \label{tbl:3}
    \resizebox{0.90\linewidth}{!}{%
        \begin{tabular}{lllll}
            \hline
             & & \textbf{Regression} & \multicolumn{2}{c}{\textbf{Classification}} \\
             \cline{3-5} \multirow{-2}{*}{\textbf{Architecture}} & \multirow{-2}{*}{\textbf{ERS}} & \boldmath m$R^{2}$ & \boldmath m$AP$ & \boldmath m$RA$ \\
             \hline
             \multicolumn{5}{l}{\cellcolor[HTML]{EFEFEF}Ablation} \\
             \hline
             $\mathcal{H} \circleddash \mathcal{H}^{\text{object}}$ & \textbf{0.3104} & \textbf{0.1286} & \textbf{0.2763} & \textbf{0.7081} \\
             $\mathcal{H} \circleddash \mathcal{H}^{\text{place}}$ & 0.2993 & 0.1233 & 0.2606 & 0.6902 \\ \hline
             \multicolumn{5}{l}{\cellcolor[HTML]{EFEFEF}Proposed architecture} \\
             \hline
             CA-MoDE & 0.3269 & 0.1404 & 0.2938 & 0.7333 \\ \hline
        \end{tabular}
    }
\end{table}

%PLACEMENT
\begin{table*}[!htbp]
    \caption{Effect of fusion strategies on prediction confidence and agreement with ground truth. Lower E indicates more confident predictions, and higher MI indicates stronger dependence between predictions and ground truth.}
    \label{tbl:4}
    \centering
    \resizebox{0.7\linewidth}{!}{%
        \begin{tabular}{lclllll}
        \hline
         &  & \multicolumn{1}{c}{\textbf{Regression}} & \multicolumn{2}{c}{\textbf{Classification}} & \multicolumn{2}{c}{\textbf{Uncertainty}} \\ \cline{3-7} 
        \multirow{-2}{*}{\textbf{Fusion method}} & \multirow{-2}{*}{\textbf{ERS}} & \boldmath m$R^{2}$ & \boldmath m$AP$ & \boldmath m$RA$ & \textbf{E~\textdownarrow} & \textbf{MI~\textuparrow} \\ \hline
        \multicolumn{7}{l}{\cellcolor[HTML]{EFEFEF}Ablation} \\ \hline
        Intermediate fusion & 0.3041 & 0.1226 & 0.2813 & 0.6902 & 1.66 & 3.70 \\
        Late fusion ($\hat{\mathbf{p}} = \mathbf{Q} \odot \mathbf{p}^{+}$) & \textbf{0.3133} & \textbf{0.1358} & 0.2715 & \textbf{0.7103} & \textbf{1.39} & \textbf{4.28} \\
        Late fusion (Kendall et al.~\cite{kendall2018multi}) & 0.3103 & 0.1291 & \textbf{0.2828} & 0.7003 & 1.51 & 3.84 \\ \hline
        \multicolumn{7}{l}{\cellcolor[HTML]{EFEFEF}Proposed architecture} \\ \hline
        CA-MoDE & 0.3269 & 0.1404 & 0.2938 & 0.7333 & 1.25 & 4.51 \\ \hline
        \end{tabular}%
    }
\end{table*}

%PLACEMENT
\begin{table*}[!htbp]
    \caption{Impact of the gating mechanism. All variants use the same CA-MoDE architecture. The comparison isolates the gating mechanism by varying only how $\mathbf{Q}$ is computed across variants.}
    \label{tbl:5}
    \centering
    \resizebox{0.71\linewidth}{!}{%
        \begin{tabular}{lclllll}
        \hline
         & & \textbf{Regression} & \multicolumn{2}{c}{\textbf{Classification}} & \multicolumn{2}{c}{\textbf{Uncertainty}} \\
         \cline{3-7} \multirow{-2}{*}{\textbf{Gating variant}} & \multirow{-2}{*}{\textbf{ERS}} & \boldmath m$R^{2}$ & \boldmath m$AP$ & \boldmath m$RA$ & \textbf{E~\textdownarrow} & \textbf{MI~\textuparrow} \\          \hline
         \multicolumn{7}{l}{\cellcolor[HTML]{EFEFEF} Ablation}
         \\          \hline
         Mean gating & 0.2994 & 0.1249 & 0.2670 & 0.6809 & 1.53 & 3.86 \\
         Learned gating & 0.3070 & 0.1303 & 0.2703 & 0.6974 & 1.37 & 4.01 \\          \hline
         \multicolumn{7}{l}{\cellcolor[HTML]{EFEFEF} Proposed architecture} \\          \hline
         CA-MoDE (Max-endorsement) & 0.3269 & 0.1404 & 0.2938 & 0.7333 & 1.25 & 4.51 \\ 
         \hline
        \end{tabular}%
    }
\end{table*}

Ablating the scene expert reduces ERS to 0.2993, while ablating the object expert reduces it to 0.3104. This gap suggests that scene context provides a stronger cue. We attribute the asymmetry to two factors. First, many BoLD clips come from older movies with lower image resolution, which inherently decreases the accuracy of $\mathcal{H}^{\text{object}}$ and propagates detection errors through the probabilistic fusion mechanism. Second, \texttt{person} dominates the detected categories in most clips, leaving most entries of $\mathbf{z}^{\text{object}}$ sparse after filtering and reducing the discriminative content of $\mathbf{y}^{\text{object}}$. 

Even so, the object expert remains useful because neither expert alone approaches the full model's ERS. The observed asymmetry also motivates our probabilistic fusion design. One expert typically dominates within a given emotion dimension, but the dominant expert can change across dimensions. For this reason, selecting the strongest contextual signal per dimension is preferable to averaging, which may dilute the leading signal with a weaker one.\newline

\noindent\textbf{(4) Contribution of Fusion Strategies:} Given that both context experts contribute independently to recognition performance, we examine how their outputs should be integrated and how the gating signal should be computed. We compare the proposed probabilistic fusion to (i) an intermediate fusion strategy, in which the outputs of $\mathcal{H}^{\text{place}}$ and $\mathcal{H}^{\text{object}}$ are concatenated with $\mathcal{H}^{\text{base}}$ features and passed through a fully connected layer before being fed to $\mathcal{H}^{\text{emotion}}$; (ii) a late fusion variant that drops the anticipated unavailable context term; and (iii) the multi-task uncertainty weighting approach proposed by Kendall et al.~\cite{kendall2018multi}.

Table~\ref{tbl:4} reports results for three fusion strategies. Intermediate fusion produces the weakest uncertainty profile because it exposes $\mathcal{H}^{\text{emotion}}$ directly to noise in the contextual outputs, a sensitivity amplified by the object detection sparsity reported in Table~\ref{tbl:3}. A late fusion variant that drops the $\mathbf{p}^{-}$ (i.e., anticipated unavailable context) term improves confidence but lowers m$AP$ compared to intermediate and late~\cite{kendall2018multi} fusion strategies. The positive co-occurrence term $\mathcal{P}^{+}$ reinforces contextually endorsed emotions with high confidence. However, when co-occurrence statistics are sparse or incorrect, those confident endorsements produce false positives that m$AP$, unlike the rank-based ERS, directly penalises. The suppression term $\mathcal{P}^{-}$ in CA-MoDE addresses this by attenuating implausible predictions. The Kendall et~al.~\cite{kendall2018multi} variant reweights losses using learned task uncertainty rather than reshaping the output distribution. While the implicit loss regularisation improves m$AP$, the ERS remains below the probabilistic late fusion strategy.\newline

\noindent\textbf{(5) Contribution of Gating Mechanism:} For gating, we substitute the proposed max-endorsement with the mean and learned variants. In all variants, the gate retains the same shifted-sigmoid form with fixed $(\alpha,\tau)$ as in Section~\ref{sec:fusion}, and only the contextual support score fed into $Q_i$ changes.

Mean gating replaces $p_{i}^{+}$ with $\bar{p}_i^{+}=\frac{1}{2}\sum_{j\in\{1,2\}}\mathcal{P}^{+}_{i,j}$ to aggregate endorsements by averaging rather than selecting the strongest signal. In learned gating, we replace $p_{i}^{+}$ with $\mathbf{w}_{i}^{\top}[\mathcal{P}^{+}_{i,1}, \mathcal{P}^{+}_{i,2}]^{\top} + b_{i}$, where $\mathbf{w}_{i} \in \mathbb{R}^{2}$ and $b_{i} \in \mathbb{R}$ are trained on BoLD.

Table~\ref{tbl:5} reports a consistent ordering across all metrics. Mean gating is the weakest variant because averaging contextual support dilutes a strong endorsement from one expert whenever the other provides no relevant cue. The lower gain of the learned gating indicates that $\mathcal{P}^{+}_{i,j}$ already encodes sufficient structure, and the additional linear layer causes overfitting rather than improving generalisation.\newline

\noindent\textbf{(6) Contribution of Facial Information:} In this experiment, we ablate facial information to quantify its contribution to CA-MoDE. To do so, we filled the face region with black pixels across all database frames and retrained the architecture with parameters described in Section~\ref{sec:architectures}. Although face masking had little effect on \texttt{person} detection in $\mathcal{H}^{\text{object}}$, ERS decreased to 0.2635 on the test set. This substantial decrease demonstrates that facial expression serves as a complementary cue that reinforces body-based emotion recognition, even in a model designed primarily to exploit contextual and postural information.

%=============SECTION VI==================
\section{Conclusion} \label{sec:conclusion}

Bodily expressions of emotion are inherently context-dependent. A raised fist in a stadium and a raised fist on a street may involve similar body configurations, but they invite different emotional interpretations. This ambiguity motivated us to propose an emotion recognition model that reasons not only about the body, but also about the contextual evidence surrounding it.

In this study, we introduced CA-MoDE, a context-aware mixture of domain experts for bodily emotion recognition in the wild. Unlike conventional approaches that incorporate scene and object information as auxiliary feature augmentations, CA-MoDE represents context through emotion-context co-occurrence priors and applies them at the output-distribution level. Scene and object experts generate soft contextual pseudo-labels, while the proposed max-endorsement gate selects the strongest contextual support for each emotion dimension. This design reduces the dilution of informative contextual cues when one expert is uncertain or uninformative.

On BoLD, CA-MoDE achieved an ERS of 0.3269 using only single still images, outperforming the compared BoLD methods that rely on temporal cues. The improvement is mainly driven by stronger categorical emotion recognition, with gains in both m$AP$ and m$RA$. At the same time, the lower VAD regression performance shows that continuous affective dimensions still require cues that static spatial context can provide only partially.

The ablation studies further support the central hypothesis of this work. Both scene and object context contribute to performance, although scene context provides the stronger signal. The max-endorsement gate outperforms mean and learned alternatives, suggesting that the co-occurrence priors already encode useful structure without requiring additional gating parameters. Finally, the face-masking experiment shows that facial information remains a complementary cue, even in a framework designed around body and context.

Overall, these findings indicate that spatial context is not merely background information for bodily emotion recognition. When modelled as structured probabilistic evidence, it can substantially improve recognition from still images and provide a complementary alternative to temporal modelling. Future work should investigate richer contextual priors, more reliable object evidence in low-resolution scenes, and hybrid models that jointly exploit spatial, facial, and temporal cues.

%=============REFERENCES==================

{\small
\bibliographystyle{IEEEtran.bst}
\bibliography{references}

@inproceedings{9666957,
  title={{Leveraging Semantic Scene Characteristics and Multi-Stream Convolutional Architectures in a Contextual Approach for Video-Based Visual Emotion Recognition in the Wild}},
  author={Pikoulis, Ioannis and Filntisis, Panagiotis P. and Maragos, Petros},
  booktitle={{2021 16th IEEE International Conference on Automatic Face and Gesture Recognition (FG 2021)}},
  pages={01--08},
  year={2021},
  doi={10.1109/FG52635.2021.9666957},
  publisher={IEEE}
}

@inproceedings{dadamo2024soniweight,
author = {D'Adamo, Amar and Roel Lesur, Marte and Turmo Vidal, Laia and Dehshibi, Mohammad Mahdi and De La Prida, Daniel and Diaz-Dur\'{a}n, Joaqu\'{\i}n R. and Azpicueta-Ruiz, Luis Antonio and V\"{a}ljam\"{a}e, Aleksander and Tajadura-Jim\'{e}nez, Ana},
title = {{SoniWeight Shoes: Investigating Effects and Personalization of a Wearable Sound Device for Altering Body Perception and Behavior}},
year = {2024},
publisher = {Association for Computing Machinery},
doi = {10.1145/3613904.3642651},
booktitle = {Proceedings of the 2024 CHI Conference on Human Factors in Computing Systems},
articleno = {93},
numpages = {20}
}

@inproceedings{li2021sequential,
  author={Li, Xinpeng and Peng, Xiaojiang and Ding, Changxing},
  booktitle={{IEEE International Joint Conference on Biometrics (IJCB)}}, 
  title={{Sequential Interactive Biased Network for Context-Aware Emotion Recognition}}, 
  year={2021},
  volume={},
  number={},
  pages={1--6},
  doi={10.1109/IJCB52358.2021.9484370},
  publisher={IEEE}
}

@inproceedings{xenos2025vllm,
  author={Xenos, Alexandros and Foteinopoulou, Niki M. and Ntinou, Ioanna and Patras, Ioannis and Tzimiropoulos, Georgios},
  booktitle={{International Joint Conference on Neural Networks (IJCNN)}}, 
  title={{VLLMs Provide Better Context for Emotion Understanding Through Common Sense Reasoning}}, 
  year={2025},
  volume={},
  number={},
  pages={1--10},
  doi={10.1109/IJCNN64981.2025.11227260},
  publisher={IEEE}
}

@inproceedings{zhang2023emotionclip,
    author={Zhang, Sitao and Pan, Yimu and Wang, James Z.},
    title={{Learning Emotion Representations From Verbal and Nonverbal Communication}},
    booktitle={{Proceedings of the IEEE/CVF Conference on Computer Vision and Pattern Recognition (CVPR)}},
    year={2023},
    pages={18993--19004},
    doi={10.1109/CVPR52729.2023.01821},
    publisher={IEEE}
}

@inproceedings{filntisis2020emotion,
  title={{Emotion Understanding in Videos Through Body, Context, and Visual-Semantic Embedding Loss}},
  author={Filntisis, Panagiotis Paraskevas and Efthymiou, Niki and Potamianos, Gerasimos and Maragos, Petros},
  booktitle={European Conference on Computer Vision Workshops},
  pages={747--755},
  year={2020},
  doi={10.1007/978-3-030-66415-2_52},
  publisher = {Springer-Verlag}
}

@article{luo2020arbee,
  title={{ARBEE: Towards Automated Recognition of Bodily Expression of Emotion in the Wild}},
  author={Luo, Y. and Ye, J. and Adams, R. B. and Li, J. and Newman, M. G. and Wang, J. Z.},
  journal={{International Journal of Computer Vision}},
  volume={128},
  number={1},
  pages={1--25},
  year={2020},
  doi={10.1007/s11263-019-01215-y},
  publisher={Springer}
}

@article{pons2020multitask,
  title={Multitask, Multilabel, and Multidomain Learning With Convolutional Networks for Emotion Recognition},
  author={Pons, G. and Masip, D.},
  journal={IEEE Trans. Cybern.},
  year={2022},
  volume={52},
  number={6},
  pages={4764--4771},
  doi={10.1109/TCYB.2020.3036935},
  publisher={IEEE}
}

@inproceedings{gu2018ava,
  title={{AVA: A Video Dataset of Spatio-Temporally Localized Atomic Visual Actions}},
  author={Gu, C. and Sun, C. and Ross, D. A. and Vondrick, C. and Pantofaru, C. and Li, Y. and Vijayanarasimhan, S. and Toderici, G. and Ricco, S. and Sukthankar, R. and Schmid, C. and Malik, J.},
  booktitle={CVPR},
  pages={6047--6056},
  year={2018},
  doi={10.1109/CVPR.2018.00633},
  publisher={IEEE}
}

@inproceedings{lin2014microsoft,
  title={{Microsoft COCO: Common Objects in Context}},
  author={Lin, T. Y. and Maire, M. and Belongie, S. and Hays, J. and Perona, P. and Ramanan, D. and Doll{\'a}r, P. and Zitnick, C. L.},
  booktitle={ECCV},
  pages={740--755},
  year={2014},
  doi={10.1007/978-3-319-10602-1_48},
  publisher={Springer Cham}
}

@inproceedings{yu2019unsupervised,
  title={{Unsupervised Person Re-Identification by Soft Multilabel Learning}},
  author={Yu, H. X. and Zheng, W. S. and Wu, A. and Guo, X. and Gong, S. and Lai, J. H.},
  booktitle={CVPR},
  pages={2143--2152},
  year={2019},
  doi={10.1109/CVPR.2019.00225},
  publisher={IEEE}
}

@inproceedings{mensink2014costa,
  title={{COSTA: Co-Occurrence Statistics for Zero-Shot Classification}},
  author={Mensink, T. and Gavves, E. and Snoek, C. G. M.},
  booktitle={CVPR},
  pages={2441--2448},
  year={2014},
  doi={10.1109/CVPR.2014.313},
  publisher={IEEE}
}

@article{zhou2018places,
  title={{Places: A 10 Million Image Database for Scene Recognition}},
  author={Zhou, B. and Lapedriza, A. and Khosla, A. and Oliva, A. and Torralba, A.},
  journal={IEEE Trans. Pattern Anal. Mach. Intell.},
  volume={40},
  number={6},
  pages={1452--1464},
  year={2018},
  doi={10.1109/TPAMI.2017.2723009},
  publisher={IEEE}
}

@inproceedings{kendall2018multi,
  title={{Multi-task Learning Using Uncertainty to Weigh Losses for Scene Geometry and Semantics}},
  author={Kendall, A. and Gal, Y. and Cipolla, R.},
  booktitle={CVPR},
  pages={7482--7491},
  year={2018},
  doi={10.1109/CVPR.2018.00781},
  publisher={IEEE}
}

@inproceedings{szegedy2015going,
  title={{Going deeper with convolutions}},
  author={Szegedy, C. and Liu, W. and Jia, Y. and Sermanet, P. and Reed, S. and Anguelov, D. and Erhan, D. and Vanhoucke, V. and Rabinovich, A.},
  booktitle={CVPR},
  pages={1--9},
  year={2015},
  doi={10.1109/CVPR.2015.7298594},
  publisher={IEEE}
}

@inproceedings{mollahosseini2016going,
  title={{Going deeper in facial expression recognition using deep neural networks}},
  author={Mollahosseini, A. and Chan, D. and Mahoor, M. H.},
  booktitle={WACV},
  pages={1--10},
  year={2016},
  doi={10.1109/WACV.2016.7477450},
  publisher={IEEE}
}

@article{kosti2020context,
  title={{Context Based Emotion Recognition Using EMOTIC Dataset}},
  author={Kosti, R. and Alvarez, J. and Recasens, A. and Lapedriza, A.},
  Journal = {IEEE Trans. Pattern Anal. Mach. Intell.},
  year = {2020},
  volume = {42},
  number = {11},
  pages={2755--2766},
  doi={10.1109/TPAMI.2019.2916866},
  publisher={IEEE}
}

@article{aviezer2012body,
  title={{Body Cues, Not Facial Expressions, Discriminate Between Intense Positive and Negative Emotions}},
  author={Aviezer, H. and Trope, Y. and Todorov, A.},
  journal={Science},
  volume={338},
  number={6111},
  pages={1225--1229},
  year={2012},
  doi = {10.1126/science.1224313},
  publisher={American Association for the Advancement of Science}
}

@article{dehshibi2021deep,
  author={Dehshibi, M. M. and Baiani, Bita and Pons, Gerard and Masip, David},
  journal={{IEEE Trans. Affect.}}, 
  title={{A Deep Multimodal Learning Approach to Perceive Basic Needs of Humans From Instagram Profile}}, 
  year={2023},
  volume={14},
  number={2},
  pages={944--956},
  doi={10.1109/TAFFC.2021.3090809},
  publisher={IEEE}
}

@article{kleinsmith2011automatic,
  title={{Automatic Recognition of Non-Acted Affective Postures}},
  author={Kleinsmith, A. and Bianchi-Berthouze, N. and Steed, A.},
  journal={IEEE Trans. Syst. Man Cybern. Part B},
  volume={41},
  number={4},
  pages={1027--1038},
  year={2011},
  doi={10.1109/TSMCB.2010.2103557},
  publisher={IEEE}
}

@article{dehshibi2023pain,
  title={{Pain Level and Pain-Related Behaviour Classification Using GRU-Based Sparsely-Connected RNNs}},
  author={Dehshibi, M. M. and Olugbade, T. A. and Diaz-de-Maria, F. and Bianchi-Berthouze, N. and Tajadura-Jim{\'e}nez, A.},
  journal={{IEEE Journal of Selected Topics in Signal Processing}},
  year={2023},
  volume={17},
  number={3},
  pages={677--688},
  doi={10.1109/JSTSP.2023.3262358},
  publisher={IEEE}
}

@article{ortigoso2025lightweight,
  author={Ortigoso-Narro, Jorge and Diaz-de-Maria, Fernando and Dehshibi, Mohammad Mahdi and Tajadura-Jim\'enez, Ana},
  journal={IEEE Sensors Journal}, 
  title={{L-SFAN: Lightweight Spatially Focused Attention Network for Pain Behavior Detection}}, 
  year={2025},
  volume={25},
  number={10},
  pages={18409--18418},
  doi={10.1109/JSEN.2025.3540415},
  publisher={IEEE}
}

@article{hens2025last,
  author={Hens, Freek and Dehshibi, Mohammad Mahdi and Bagheriye, Leila and Tajadura-Jiménez, Ana and Shahsavari, Mahyar},
  journal={IEEE Transactions on Neural Systems and Rehabilitation Engineering}, 
  title={{LAST-PAIN: Learning Adaptive Spike Thresholds for Low Back Pain Biosignals Classification}}, 
  year={2025},
  volume={33},
  number={},
  pages={1038--1047},
  doi={10.1109/TNSRE.2025.3546682},
  publisher={IEEE}
}

@article{fang2024eva,
title = {{EVA-02: A visual representation for neon genesis}},
journal = {Image and Vision Computing},
volume = {149},
pages = {105171},
year = {2024},
issn = {0262-8856},
doi = {10.1016/j.imavis.2024.105171},
author = {Yuxin Fang and Quan Sun and Xinggang Wang and Tiejun Huang and Xinlong Wang and Yue Cao},
publisher = {Elsevier}
}

@INPROCEEDINGS{he2016deep,
  author={He, Kaiming and Zhang, Xiangyu and Ren, Shaoqing and Sun, Jian},
  booktitle={2016 IEEE Conference on Computer Vision and Pattern Recognition (CVPR)}, 
  title={Deep Residual Learning for Image Recognition}, 
  year={2016},
  volume={},
  number={},
  pages={770--778},
  doi={10.1109/CVPR.2016.90},
  publisher={IEEE}
}
}

\end{document}